\documentclass{article}

\usepackage[final]{neurips_2025}

\usepackage[utf8]{inputenc} 
\usepackage[T1]{fontenc}    
\usepackage[hidelinks]{hyperref}       
\usepackage{url}            
\usepackage{booktabs}       
\usepackage{amsfonts}       
\usepackage{nicefrac}       
\usepackage{microtype}      
\usepackage{xcolor}         

\usepackage{amsmath,amsfonts,bm}

\def\Wenc{\mathbf{W}_\mathrm{enc}}
\def\Wdec{\mathbf{W}_\mathrm{dec}}
\def\Wdeci{\mathbf{W}_\mathrm{dec}^{(i)}}
\def\Wenci{\mathbf{W}_\mathrm{enc}^{(i)}}
\def\Wencj{\mathbf{W}_\mathrm{enc}^{(j)}}
\def\Wdecj{\mathbf{W}_\mathrm{dec}^{(j)}}
\def\WU{\mathbf{W}_\mathrm{U}}

\def\eqref#1{equation~\ref{#1}}

\def\1{\bm{1}}

\DeclareMathAlphabet{\mathsfit}{\encodingdefault}{\sfdefault}{m}{sl}
\SetMathAlphabet{\mathsfit}{bold}{\encodingdefault}{\sfdefault}{bx}{n}

\DeclareMathOperator*{\argmax}{arg\,max}

\let\cite\citep

\usepackage[textsize=scriptsize]{todonotes}
\usepackage{xspace}

\makeatletter
\newcommand*\iftodonotes{\if@todonotes@disabled\expandafter\@secondoftwo\else\expandafter\@firstoftwo\fi}  
\makeatother

\definecolor{darkblack}{rgb}{0.0,0.0,0.5}
\definecolor{darkgreen}{rgb}{0.0, 0.42, 0.24}
\definecolor{lightgreen}{rgb}{0.52, 0.73, 0.4}
\definecolor{darkgray}{rgb}{0.4,0.4,0.4}
\definecolor{darkblue}{rgb}{0.0,0.0,0.5}
\definecolor{darkpurple}{rgb}{0.5,0.2,0.8}
\definecolor{lightpurple}{rgb}{0.8,0.5,1}

\usepackage{cleveref}
\crefname{figure}{Figure}{Figures}
\crefname{table}{Table}{Tables}
\crefname{appendix}{Appendix}{Apps.}
\crefname{section}{\S}{\S\S}
\crefformat{section}{\S#2#1#3} 
\crefname{equation}{Eq.}{Eqs.}
\crefname{algorithm}{Alg.}{Algs.}
\crefname{algocf}{Alg.}{Algs.}

\usepackage{wrapfig}
\usepackage{tabularx}
\usepackage{booktabs}
\usepackage{dsfont}
\usepackage{multirow}
\usepackage{array}

\usepackage{caption}
\usepackage{listings}
\usepackage{adjustbox}

\usepackage{float}
\usepackage{enumitem}

\title{Dense SAE Latents Are Features, Not Bugs}

\author{%
  Xiaoqing Sun\thanks{Equal contribution. Correspondence to \texttt{xqsun@mit.edu} and \texttt{stolfoa@ethz.ch}.} \\
   MIT \\
   \And
   Alessandro Stolfo$^*$\\
     ETH Z\"urich 
  \AND
    Joshua Engels \\
   MIT 
   \And
   Ben Wu  \\
   University of Sheffield \\
   \And
   Senthooran Rajamanoharan \\
    \\
   \AND
   Mrinmaya Sachan \\
   ETH Z\"urich \\
   \And
   Max Tegmark \\
   MIT \\
}

\begin{document}

\maketitle

\begin{abstract}
  Sparse autoencoders (SAEs) are designed to extract interpretable features from language models by enforcing a sparsity constraint. Ideally, training an SAE would yield latents that are both sparse and semantically meaningful. However, many SAE latents activate frequently (i.e., are \emph{dense}), raising concerns that they may be undesirable artifacts of the training procedure. In this work, we systematically investigate the geometry, function, and origin of dense latents and show that they are not only persistent but often reflect meaningful model representations. We first demonstrate that dense latents tend to form antipodal pairs that reconstruct specific directions in the residual stream, and that ablating their subspace suppresses the emergence of new dense features in retrained SAEs---suggesting that high density features are an intrinsic property of the residual space. We then introduce a taxonomy of dense latents, identifying classes tied to position tracking, context binding, entropy regulation, letter-specific output signals, part-of-speech, and principal component reconstruction. Finally, we analyze how these features evolve across layers, revealing a shift from structural features in early layers, to semantic features in mid layers, and finally to output-oriented signals in the last layers of the model. Our findings indicate that dense latents serve functional roles in language model computation and should not be dismissed as training noise.
\end{abstract}

\section{Introduction}
Sparse autoencoders (SAEs) are an unsupervised method for extracting interpretable features from language models \cite{bricken2023monosemanticity, huben2024sparse, kissane2024interpreting}. They address the challenge of polysemanticity, where individual neurons activate in semantically diverse contexts that defy a single explanation \cite{olah2017feature, elhage2022superposition}. SAEs are trained to reconstruct the activations of a language model under a sparsity constraint applied to a bottleneck layer, ensuring that only a small subset of latents are active at a time.\footnote{We use ``latent'' to refer to an entry in the SAE’s sparse hidden layer.} This method effectively recovers interpretable features in a variety of models, including Claude 3 Sonnet \cite{templeton2024scaling} and GPT-4 \cite{gao2025scaling}.

Ideally, a trained SAE would yield a large set of interpretable and sparsely activating latents. In practice, however, SAEs exhibit a substantial fraction of densely activating latents, activating on 10\% to 50\% of tokens \cite{topkvsgated, rajamanoharan2024jumpingaheadimprovingreconstruction}. These dense latents are challenging to interpret based solely on their activation patterns. It remains unclear whether they arise as an optimization by-product, or if they instead capture inherently dense signals present in the model’s residual stream \cite{anthropic-interpretable-dense-features, removing-dense-latents}.

In this work, we investigate several properties of dense SAE latents and the residual stream subspaces they span, uncovering evidence that these latents track meaningful residual stream information. First, we observe that when retraining an SAE on model activations with the dense latent space ablated, virtually no dense latents are learned---dense latents reflect an intrinsic property of the residual stream rather than a training artifact. We then study the geometry of dense latents and observe that they tend to form antipodal pairs, with each pair effectively reconstructing a single direction.

We then examine the Gemma Scope suite of SAEs \cite{lieberum2024gemmascopeopensparse} across layers to propose a taxonomy of dense latents.
We identify latents whose activations encode positional information, latents reconstructing a subspace of the residual stream linked to entropy regulation \cite{stolfo2024confidence, cancedda-2024-spectral}, latents tracking high-level shifts in the text, latents encoding letter-specific output signals, latents tracking parts of speech, and latents reconstructing the first residual stream principal component direction. We additionally examine how these dense latents transform across layers, finding that there is a pronounced increase in the number of dense latents just before the unembedding, as well as a shift from structural signals in early layers (e.g., position tracking) to output‐oriented signals at the end. Our findings provide evidence that dense SAE latents reflect inherently dense mechanistic functions within language models. 

\section{Background}
\paragraph{SAEs.} Sparse autoencoders (SAEs) are trained to reconstruct a language model’s activations $\mathbf{x} \in \mathbb{R}^{d_{\mathrm{model}}}$ while imposing a sparsity constraint \cite{yun-etal-2021-transformer, huben2024sparse}. This computation can be represented as:
\begin{align*}
    \mathbf{f}(\mathbf{x}) & := \sigma(\mathbf{W}_\mathrm{enc} \mathbf{x} + \mathbf{b}_\mathrm{enc}),\\
    \hat{\mathbf{x}}(\mathbf{f}) &:= \mathbf{W}_\mathrm{dec} \mathbf{f} + \mathbf{b}_\mathrm{dec},
\end{align*}
where $ \mathbf{f}(\mathbf{x}) \in \mathbb{R}^{d_\mathrm{sae}}$ is a sparse, non-negative vector of latents, with $d_\mathrm{sae} \gg d_\mathrm{model}$, and $\sigma$ is a non-linear activation function. SAEs are typically trained to minimize the L2 distance between the original activation and its reconstruction $\|\mathbf{x} - \hat{\mathbf{x}}(\mathbf{f}(\mathbf{x})) \|_2^2$ while a sparsity constraint is imposed on $\mathbf{f}$ by adding a sparsity-related loss component or via specific activation functions.
We denote the encoder and decoder weights of the latent at index $i$ as $\Wenci$ and $\Wdeci$, respectively. Unless noted otherwise, we use ``dense'' to refer to latents with an activation frequency larger than 0.1.

\paragraph{Experimental Setup.}
We focus our investigation on the Gemma Scope SAEs \cite{lieberum2024gemmascopeopensparse} trained on Gemma 2 2B \cite{gemmateam2024gemma2improvingopen}, which use a JumpReLU activation function \cite{rajamanoharan2024jumpingaheadimprovingreconstruction}. We additionally train TopK SAEs \cite{gao2025scaling} on 1B tokens of the OpenWebText corpus \cite{gokaslan2019openweb} for our experiments in \cref{sec:dense_ablation}.\footnote{We choose TopK for its reliable training and competitive reconstruction–sparsity trade-off.} Activation densities for Gemma Scope latents are from Neuronpedia \cite{neuronpedia}, while densities for our TopK SAEs are computed over 100M tokens from the C4 Corpus \cite{c4}. Full experimental details are in \cref{app:exp_details}.

\section{General Properties of Dense Latents}
We begin by examining structural properties of dense SAE latents, finding that they arise from a specific residual stream subspace (\cref{sec:dense_ablation}), and that they tend to cluster in antipodal pairs (\cref{sec:antipodality}).

\subsection{Dense Latents Reflect Intrinsic Properties of the Residual Stream}
\label{sec:dense_ablation}
To determine whether dense SAE latents arise from the training procedure or reflect an intrinsic property of the residual‐stream subspace they reconstruct, we perform a targeted ablation experiment. We identify the subspace spanned by the dense latents of an SAE trained on layer 25 of Gemma 2 2B, then train a new SAE on activations in which this subspace has been zero-ablated. For comparison, we also select an equally sized set of non‐dense latents and train a third SAE after ablating their subspace.  We repeat this for two dictionary sizes ($d_{\mathrm{sae}}=16384$ and $32768$).

\begin{figure}[t]
    \centering
    \includegraphics[width=0.325\textwidth]{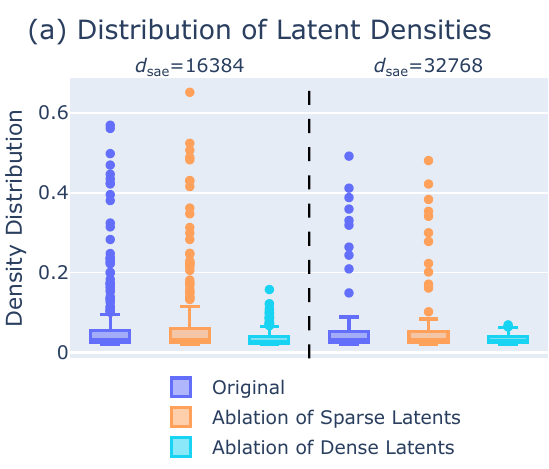}
    \includegraphics[width=0.325\textwidth]{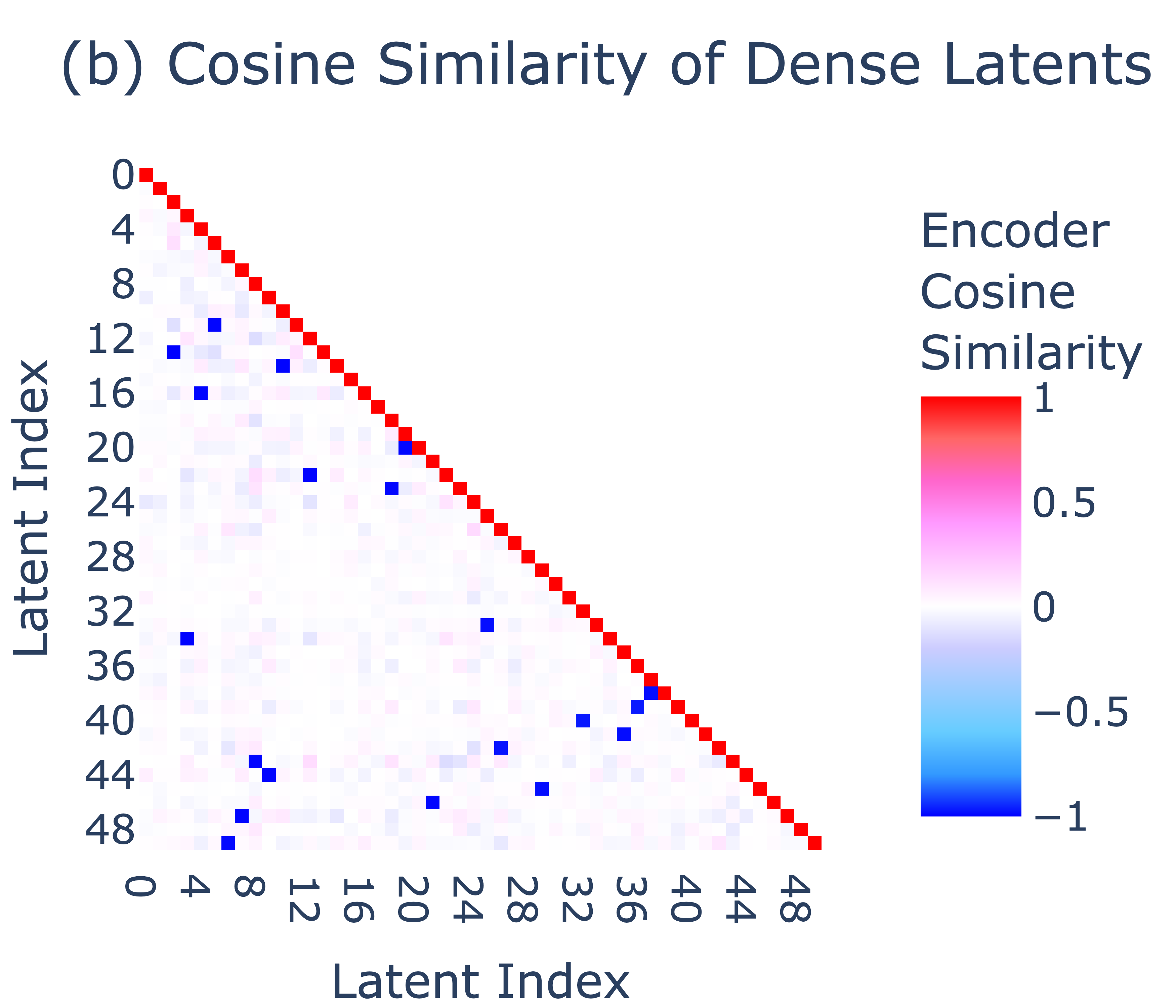}
    \includegraphics[width=0.325\textwidth]{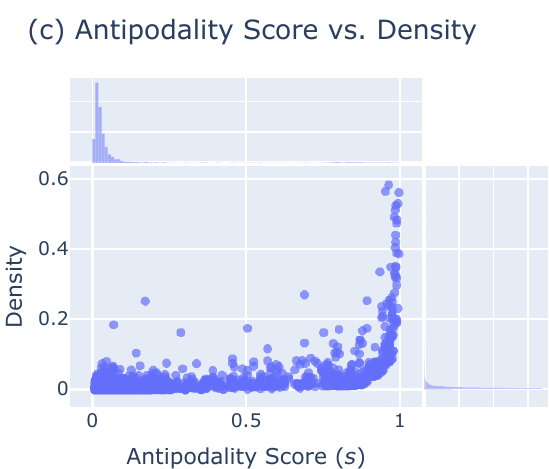}
    
\caption{\textbf{General Properties of Dense SAE Latents.} 
(a) Ablating the dense-latent subspace (teal) reduces high-density latents compared to the original (blue) and sparse-latent ablations (orange).
(b) Encoder cosine similarity between the top 50 latents with highest density.
(c) Dense latents exhibit high antipodality score: they form pairs that reconstruct specific residual stream directions.}
  \label{fig:general_characteristics}
  \vspace{-3mm}
\end{figure}

\cref{fig:general_characteristics}a shows the resulting distributions of latent activation densities. In both dictionary sizes, ablating the dense‐latent subspace (teal) yields much fewer high‐density latents than the original SAE (blue) and the non‐dense ablation (orange). This result implies that densely activating latents are not mere training artifacts but instead track a dense residual‐stream subspace whose presence drives the emergence of dense latents. As additional evidence that dense latents are not training artifacts, in \cref{app:training} we show that longer training does not reduce the number of dense latents. We further replicate this dense‐subspace ablation on GPT-2 and LLaMA 3.2 with the same outcome (\cref{app:dense_ablations}).\looseness=-1


\subsection{Dense Latents Cluster in Antipodal Pairs}
\label{sec:antipodality}

We now examine the geometry of dense latents and observe that they tend to form antipodal pairs. That is, as shown in \cref{fig:general_characteristics}b, there exist many pairs of two dense latents that have nearly opposite decoder vectors (we find a similar result for encoder vectors). This suggests that the SAE allocates two latents in the dictionary to represent a 1-dimensional line. 


To quantify whether this phenomenon is specific to dense latents, we introduce an antipodality score $s_i$ for a latent $i$. We first compute the pairwise cosine similarities between the latent's weights (both encoder and decoder) and those of all other latents. Then, we compute the maximum product of encoder and decoder cosine similarity across all pairs $(i, j)$ for all $i \neq j$. Formally, we have
\begin{equation}\label{eq:antipodality_score}
s_i := \max_{j \neq i} \left( \operatorname{sim}\big(\Wenci, \Wencj\big) \cdot \operatorname{sim}\big(\Wdeci, \Wdecj\big) \right),
\end{equation}

\begin{wrapfigure}{r}{0.5\textwidth} 
  \vspace{-\intextsep} 
  \centering
  \includegraphics[width=\linewidth]{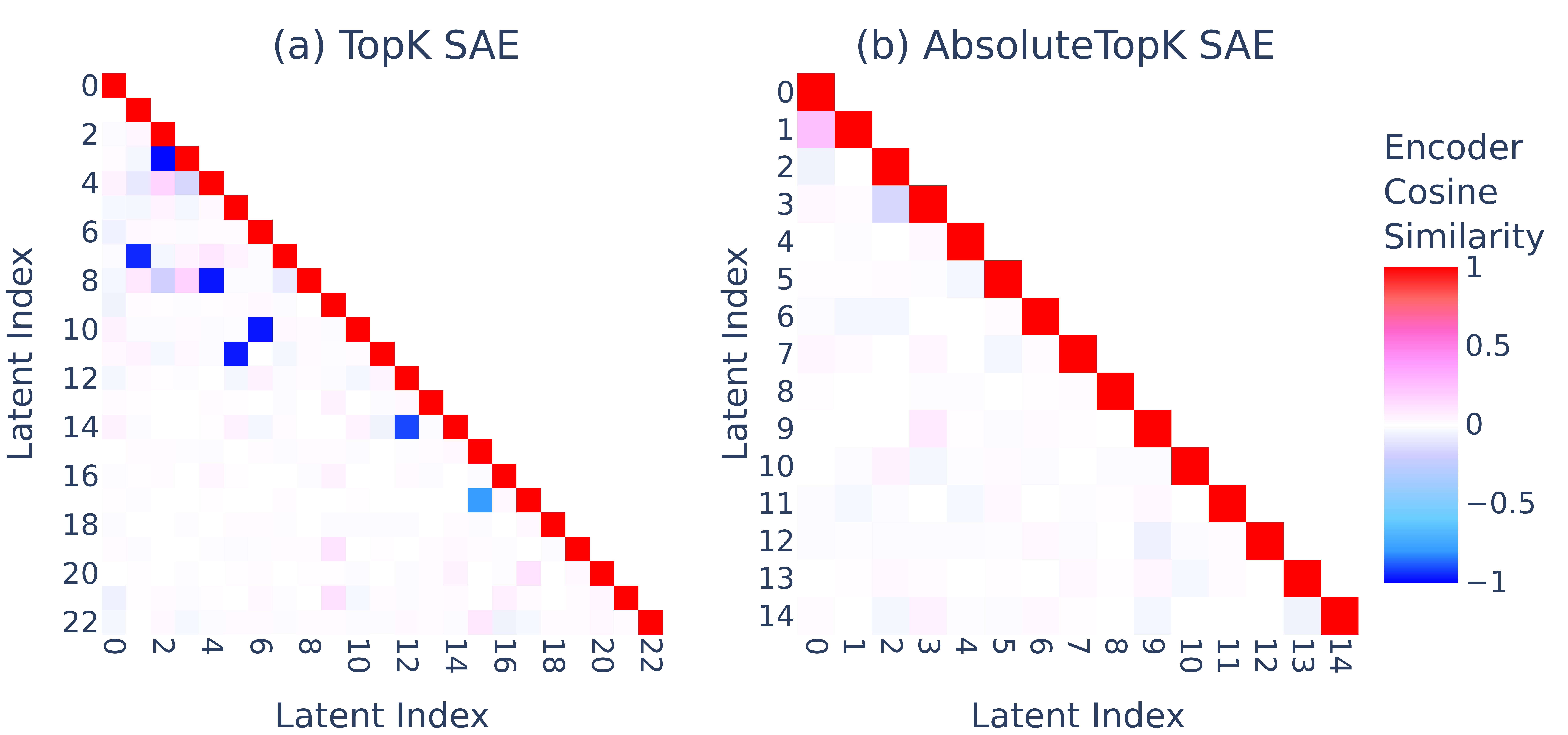}
  \caption{\textbf{AbsoluteTopK SAEs show no antipodality.} Allowing the SAE to have both positive and negative latent activations removes antipodal dense latents.}
  \vspace{-\intextsep}
  \label{fig:abs}
\end{wrapfigure}

where $\operatorname{sim}(u,v)$ denotes the cosine similarity between vectors $u$ and $v$.
This score reflects the extent to which latent $i$ forms an antipodal pairing with another latent: high values of $s_i$ indicate that there is another latent $j$ with both encoder and decoder weights nearly opposite in direction to those of $i$.\footnote{Although high values of $s$ could be produced by two nearly identical latents, retaining such a pair would be redundant--a scenario we do not observe. Evidence for this is provided in \cref{app:cosine_sim}.}

As shown in \cref{fig:general_characteristics}c, $s_i$ and the activation density of latent $i$ are strongly positively correlated. The majority of dense latents---particularly those with an activation frequency exceeding 0.3---exhibit pairwise scores greater than 0.9, supporting our conclusions above. We provide density-antipodality visualizations for additional SAEs in \cref{app:antipodality}, showing that this trend holds consistently across SAE architectures (JumpReLU and TopK), models (GPT-2 and Gemma), and layers. 

Additionally, we train an AbsoluteTopK SAE, which allows activations of SAE latents to be negative, and enforces sparsity by taking the TopK latents with greatest absolute activations. This effectively allows the same latent direction to be used in both ``positive'' and ``negative'' directions for reconstruction. We compare this to a TopK SAE trained with the same seed, and show that this AbsoluteTopK activation function eliminates the antipodal dense latents (Figure \ref{fig:abs}).

\section{Taxonomy}
\label{sec:taxonomy}
Having established that dense latents are persistent and geometrically structured, we now investigate their interpretability. We identify classes of dense latents based on the model signals they represent:

\begin{itemize}[align=left,leftmargin=*]
    \item \textbf{Position latents} (\cref{sec:position_latents}) fire based on token position relative to structural boundaries (start of sentence, paragraph or context) and appear early in the network.
    \item \textbf{Context-binding latents} (\cref{sec:discourse_latents}) represent context-dependent semantic content and exhibit coherent chunk-level activations, potentially representing high-level ideas within the context.
    \item \textbf{Nullspace latents} (\cref{sec:nullspace}) track components of the residual stream that have minimal impact on next token prediction. They instead regulate prediction entropy.
    \item \textbf{Alphabet latents} (\cref{sec:alphabet}) promote broad sets of tokens sharing an initial character.
    \item \textbf{Meaningful-word latents} (\cref{sec:noun_latents}) have activations related to the token part-of-speech tag.
    \item \textbf{PCA latents} (\cref{sec:pca_latents}) lie almost completely within the first PCA components of the activation space.
\end{itemize}

\begin{figure}
    \centering
    \includegraphics[width=1\linewidth]{figures/master.pdf}
    \caption{\textbf{An overview of our taxonomy of dense latents, for every layer.} See \cref{app:class} for how we created this plot.}
    \label{fig:master_taxonomy}
    \vspace{-3mm}
\end{figure}

\subsection{Position Latents}

\label{sec:position_latents}
We first identify a class of dense latents whose activations track the current token’s position relative to specific text boundaries. \textbf{Context-tracking} latents track token position w.r.t. the BOS token, \textbf{paragraph-tracking} latents track token position w.r.t. a paragraph start, and \textbf{sentence-tracking} latents track token position w.r.t. a sentence beginning. Context-position latents are similar to ``position neurons'' from prior work \citep{gurnee2024universalneuronsgpt2language}; the other categories are to the best of our knowledge novel.

To find these latents systematically, we use Spearman's rank correlation coefficient $\rho$. For each dense latent, we capture the projections\footnote{We use the projection of the residual stream rather than the JumpReLU activations of these latents since we hypothesize that the \textit{direction} itself encodes the positional information, regardless of whether the magnitude exceeds the learned JumpReLU threshold.} of the residual stream activations onto its decoder vector for 5000 1024-token-long contexts. We find $\rho$ between this projection and the distance from the last period, the last newline, and the beginning of the input. These boundaries act as proxies for ``beginning of sentence'', ``beginning of paragraph'', and ``beginning of context'', respectively. 

\cref{fig:master_taxonomy} shows the resulting trends: sentence-tracking and paragraph-tracking latents are prominent before layer 10, while context-position tracking latents are present throughout the model. \cref{fig:position} shows $\rho$ for all latents across layers. We can clearly see groups of outlier latents for each category, and thus classify latents as belonging to that category if $|\rho| > 0.4$. Indeed, examples in \cref{app:position} confirm that the identified outlier latents have position-tracking behavior. Notably, \cref{app:position} also shows that paragraph-tracking latents are agnostic to artificially adding formatting newlines, suggesting that this direction in the model tracks true semantic paragraph breaks. Thus, our ``distance to newline objective'' is just a proxy. We also note that latents with high $\rho$ with periods also have high $\rho$ with newlines, since newlines and periods are correlated in text. In \cref{fig:newline_period}, we thus show the $\rho$ for sentence-tracking vs. paragraph-tracking across all dense latents.

At a higher level, it makes sense that the model represents these features in a dense way: positional information is always relevant to the model's predictions (e.g., it must track how far it is in a sentence to correctly predict a period), so the model might store this representation in a consistent direction in every hidden state, which is then learned by the SAE.


\subsection{Context-Binding Latents}
\label{sec:discourse_latents}
\begin{figure}
    \centering
    \includegraphics[width=1\linewidth]{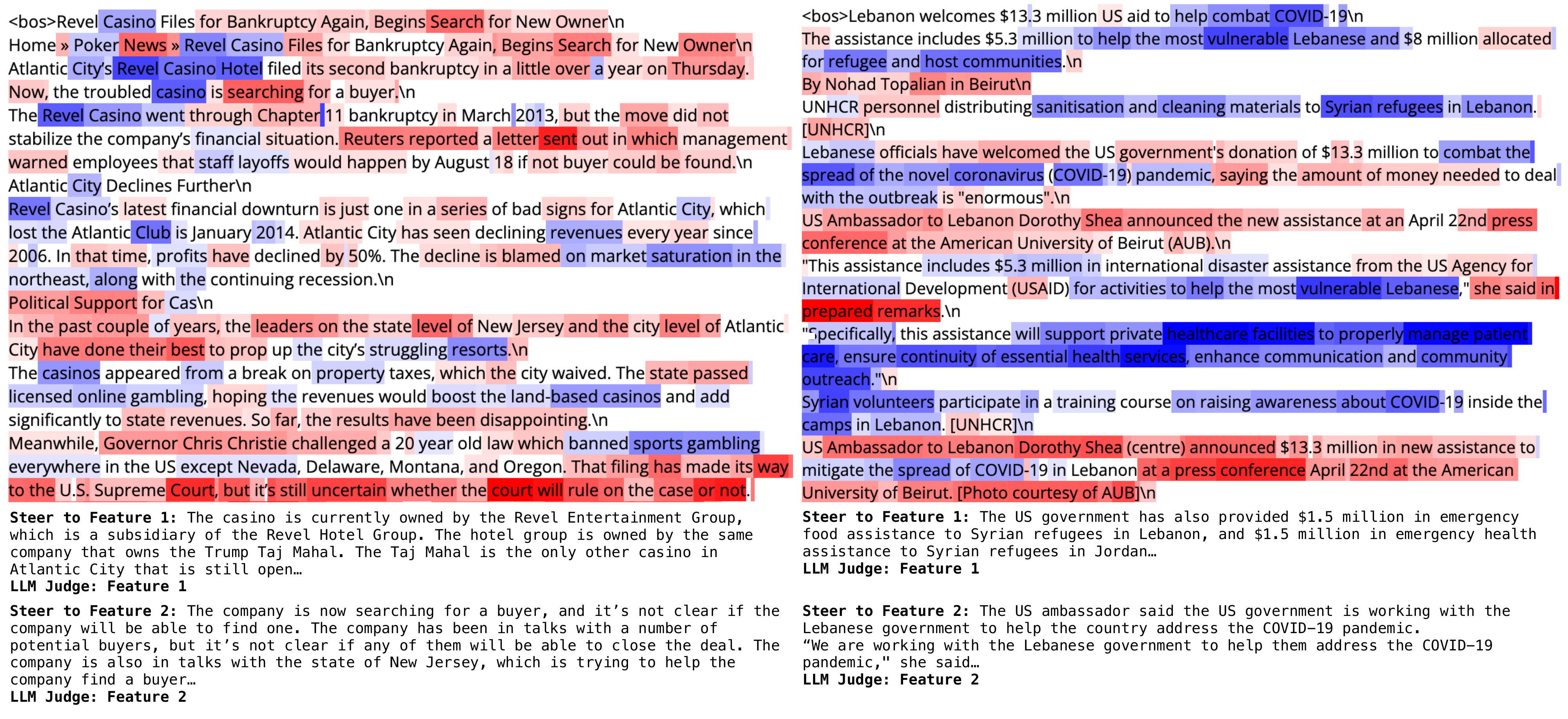}
    \caption{\textbf{Context-Binding Latents.} Activation patterns of layer 12 antipodal pair 7541 (blue, feature 1) and 2009 (red, feature 2). In the first context, they seem to be tracking ``casino facts'' vs ``looking for a buyer'', while in the second context, they seem to be tracking ``healthcare'' vs ``press conference''. Their corresponding completions are in line with the concepts they activated on.}
    \label{fig:context-binding}
\end{figure}


We next identify a class of dense latents that encode different semantic concepts depending on context. Unlike interpretable sparse SAE latents typically associated with fixed meanings, such as the “Golden Gate Bridge” feature in Claude \citep{templeton2024scaling}, these dense latents appear to \textit{bind} to the main ideas of the context.

We first observe that some dense latents, particularly in middle layers, activate on long consecutive ``chunks'' of tokens.\footnote{While positional latents also exhibit consecutive activations, here we refer to non-positional latents whose activations cannot be explained by position alone.} Examining the activations of such latents, we notice empirically that such latents fire on highly specific concepts \textit{within a context}, but the concepts \textit{vary across contexts}. We generate explanations of these latents with an LLM and confirm that they seem to be more context-specific than sparse latents (see \cref{app:autointerp}).
 
One possible interpretation is that these latents represent general but abstract, difficult-to-interpret properties. However, we also observe that within an antipodal pair, the active latent often switches when the main topic or entity in the text changes (\cref{fig:context-binding}, \cref{app:context_binding_examples}). This raises the hypothesis that such directions act as ``registers'' in the residual stream for tracking the active concept, rather than simply representing generic properties. 

We thus perform a steering experiment to find the causal effect of these directions. For each antipodal pair (F1, F2), we prompt Gemma 2 2B with input text from the RedPajama dataset \citep{weber2024redpajama} and generate completions without steering, steering to F1, and steering to F2. An LLM judge \citep{google2025gemini} is then asked whether each completion is more in line with activating examples (from the input context) of F1 or F2, or unclear. Further details of the methodology are in \cref{app:steering_context_binding}.

Since the unsteered generation may already favor F1 or F2, we quantify steering success by the fraction of \textit{flips} from the unsteered judgment that align correctly with the steering direction. For several mid-layer latent pairs, steering reliably shifts completions towards the specific concept previously associated with the latent \textit{in that context} (\cref{fig:flips}). However, when judged against out-of-context examples, the rate of unclear judgments rises sharply (\cref{tab:unknown}). While difficult to rule out the possibility that these directions encode ``general uninterpretable'' features, the specificity of the steered generation in bringing up context-related ideas suggests that these latents could bind to concepts in a context-dependent, rather than globally consistent, way. \looseness=-1

Previous works have uncovered ``binding mechanisms'' that help the model keep track of in-context associations between entities \citep{jiahai_binding, feng2024monitoringlatentworldstates}.
While our findings do not directly prove such a mechanism, they raise the possibility that dense subspaces may play a similar functional role, distinguishing the currently active semantic concept. Further work could explore the circuits \citep{marks2025sparse} involving such subspaces, and challenge the assumption of globally monosemantic directions.

\begin{figure}[t]
\centering
\begin{minipage}{0.35\textwidth}
    \centering
    \includegraphics[width=\linewidth]{figures/flips.pdf}
    \caption{\textbf{Fraction of correct flips when steering}, for all latent pairs that have at least one latent $f>0.2$, and $\geq40$ flips. Points are sized by number of flips.}
    \label{fig:flips}
\end{minipage}
\hfill
\begin{minipage}{0.6\textwidth}
\centering
\footnotesize
\begin{tabular}{|c|c|c|c|}
\hline
\textbf{Layer} & \textbf{Latent Pair} & \textbf{In-context} & \textbf{Out-of-context} \\
\hline
12 & (14906, 14599) & 0.051 & 0.717 \\
12 & (2291, 13295) & 0.028 & 0.760 \\
12 & (7541, 2009) & 0.043 & 0.711 \\
13 & (3517, 46) & 0.036 & 0.742 \\
13 & (15275, 11449) & 0.029 & 0.704 \\
13 & (12613, 7655) & 0.028 & 0.531 \\
14 & (11575, 2411) & 0.047 & 0.798 \\
14 & (8515, 15297) & 0.041 & 0.603 \\
14 & (6699, 1802) & 0.037 & 0.678 \\
16 & (2889, 8811) & 0.024 & 0.665 \\
17 & (10495, 491) & 0.051 & 0.669 \\
\hline
\end{tabular}
\captionof{table}{\textbf{Fraction of ``unclear'' judgments} using in-context examples versus out-of-context examples, for the highest-scoring latents by flips.}
\label{tab:unknown}
\end{minipage}
\vspace{-3mm}
\end{figure}

\subsection{Nullspace Latents}

\label{sec:nullspace}
Previous work has identified a $\WU$ \emph{quasi-nullspace}--the subspace spanned by the last singular vectors of the unembedding matrix $\WU$--which accounts for a substantial portion of the residual stream’s norm, yet has little direct impact on next‐token prediction \cite{cancedda-2024-spectral}. Since this subspace carries high norm, we hypothesize that some dense SAE latents are allocated specifically to reconstruct it.

\begin{figure}[t]
    \centering
    \includegraphics[width=0.325\textwidth]{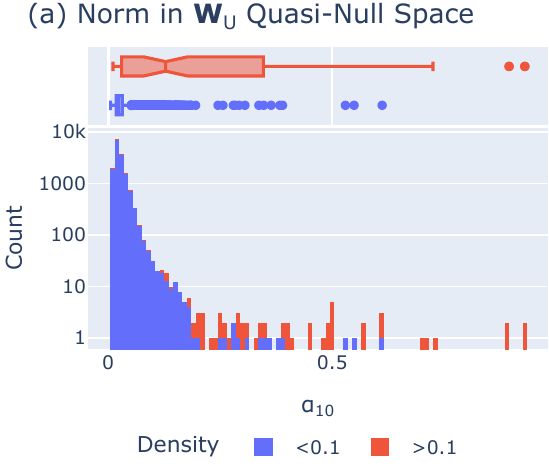}
    \includegraphics[width=0.325\textwidth]{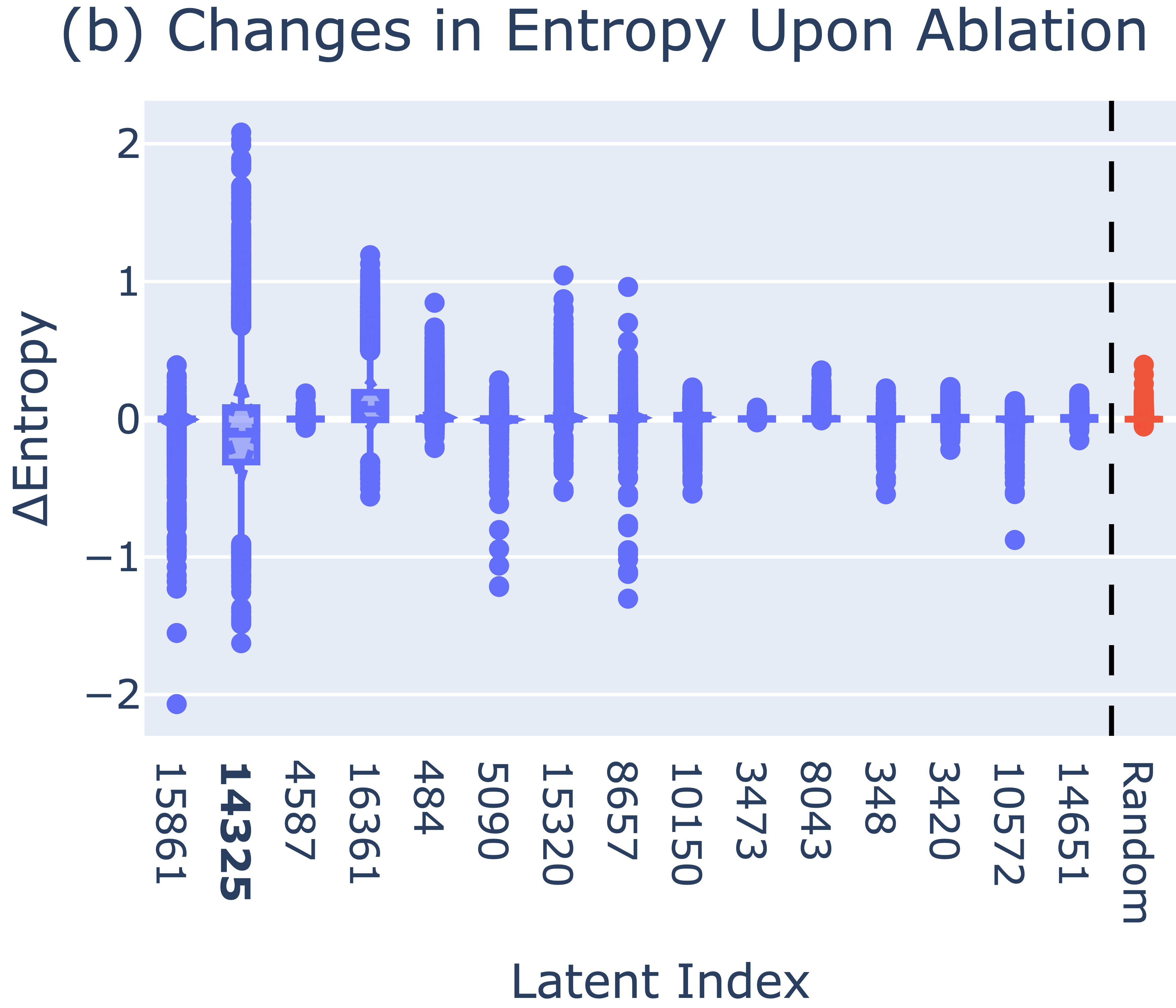}
    \includegraphics[width=0.325\textwidth]{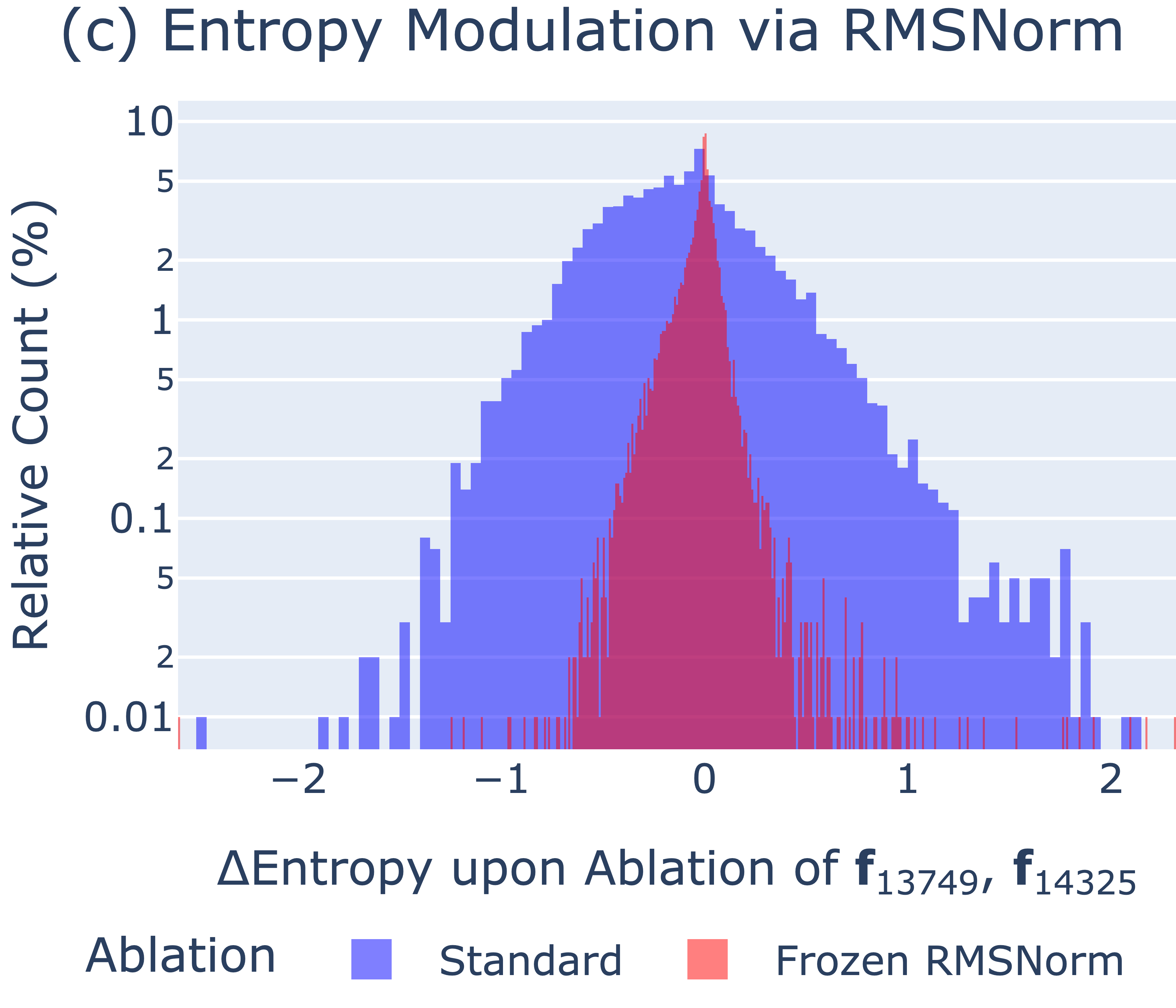}
\caption{\textbf{Nullspace Latents.} 
(a) A small fraction of latents concentrate norm in the final 10 singular directions of $\WU$, with high-density latents overrepresented in this group. (b) A pair of such latents correlates strongly with model output entropy. (c) Ablating this pair lowers entropy; the effect substantially decreases when RMSNorm scaling is frozen.}
  \label{fig:nullspace_latents}
\end{figure}

To test this, we compute the singular value decomposition $\WU = \mathbf{U\Sigma V}^\mathrm{T}$. Then, we study the composition of an SAE latent $i$’s encoder weight with the space spanned by the last $k$ left singular vectors $\mathbf{U}_{-k}, \dots, \mathbf{U}_{-1}$ of $\WU$ by computing the fraction $\rho_k$ of the norm of its encoder weight $\Wenci$ that lies in this subspace:
\begin{equation}\label{eq:fract_of_norm}
\alpha_k = \frac{\sum_{j=1}^{k} \mathbf{U}_{-j}^\mathrm{T} \Wenci}{\| \Wenci \|}.
\end{equation}
A histogram of $\alpha_{10}$ for the SAE trained at layer 25 of Gemma 2 2B (\cref{fig:nullspace_latents}a) shows that $99.6\%$ of latents have $\alpha_{10}<0.2$. We designate those with $\alpha_{10}>0.2$ as \emph{nullspace‐aligned}. Interestingly, $75\%$ of them are high-density, and account for $40\%$ of the high-density latents in the SAE.

Unlike other dense latents, nullspace‐aligned latents are hard to interpret via their token‐level activation patterns. Additionally, the tokens they promote are typically uninterpretable ``under‐trained'' tokens \cite{land-bartolo-2024-fishing}. Motivated by prior work linking the $\WU$ nullspace to an RMSNorm-based \cite{rmsnorm} entropy regulation mechanism \cite{stolfo2024confidence}, we investigate whether these latents encode this internal computation.

To test whether these latents causally influence output entropy, we ablate the residual stream along each latent’s decoder direction by setting its value to the corresponding decoder bias, thereby removing information in that direction. We then measure the change in per-token entropy of the model’s output distribution. \cref{fig:nullspace_latents}b reports the entropy change for all latents with $\alpha_{10} > 0.3$ (one per antipodal pair to avoid redundancy), compared to a control group of 50 randomly selected latents.\footnote{The entropy changes for the random latents are aggregated into a single boxplot.} 

\begin{figure}
  \begin{minipage}[b]{.32\textwidth}
\centering
    \includegraphics[width=1\textwidth]{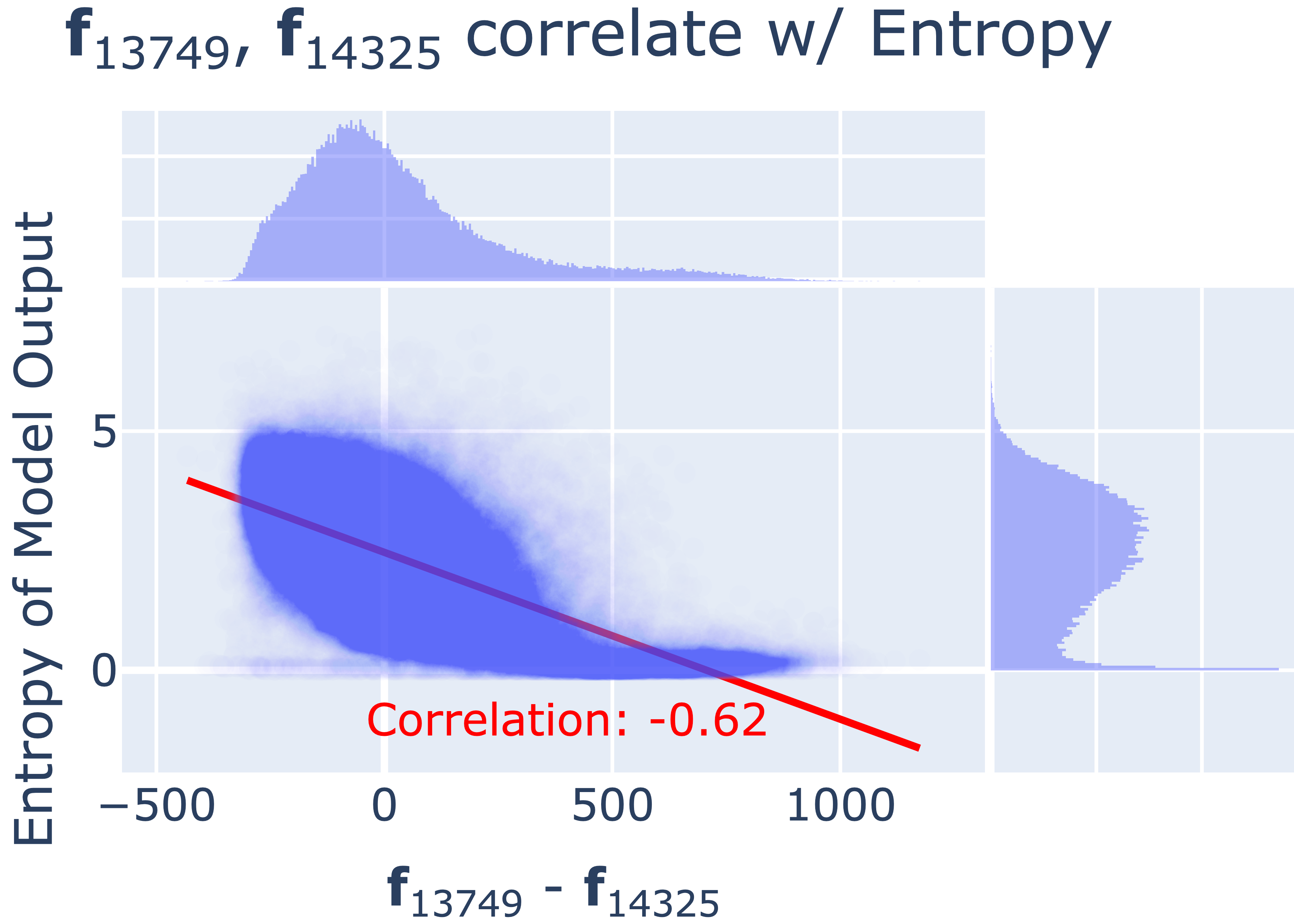}
    \captionof{figure}{\textbf{Entropy Correlation.} A pair $\WU$ nullspace-aligned correlates strongly with model output entropy.}
  \label{fig:entropy_correlation}
  \end{minipage}\hfill
 \begin{minipage}[b]{.65\textwidth}
    \resizebox{1\textwidth}{!}{
    \begin{tabularx}{1.45\textwidth}{l c c c l }
    \toprule[0.1em]
     \textbf{Index} & \textbf{Letter} & \textbf{Density} & \textbf{Metric} & \textbf{Top Tokens}  \\
        \midrule
    15287 & R & 0.16 & 0.98 &  \texttt{\_RI, \_rb, getR, \_ri, \_r, \_RS, R, \_RR}\\
    13531 & M & 0.15 & 0.97 &  \texttt{\_MM, \_m, MM, \_mM, \_mm, \_mf, \_ms, mM}\\
    30 & T & 0.16 & 0.99 &  \texttt{\_TT, \_TC, TT, TC, \_tc, \_TG, \_TS, \_TD}\\
    1761 & D & 0.14 & 0.98 & \texttt{\_DD, \_D, \_DS, \_DP, \_DT, DD, DP, DS, \_Ds}\\
    7342 & I & 0.13 & 0.91 & \texttt{IB, i, IC, İ, IE, IH, IP, \_IW, IR, IW}\\
    2651 & U & 0.11 & 0.93 & \texttt{\_UA, U, \_UT, UU, \_U, \_UF, \_UD, UE, UA}\\
    4664 & C & 0.14 & 0.93 & \texttt{\_getC, \_CC, getC, \_c, setC, CC, Cs, \_Cs}\\
    \midrule
    357 & B(+R) & 0.006 & 0.91 & \texttt{\_BR, \_Br, Br, BR, \_Bra, \_br, Bra, br}\\
    12114 & S(+L) & 0.006 & 0.95 & \texttt{\_SL, SL, \_sl, \_Sl, sl, Sl, \_Slide}\\
    14857 & C(+U) & 0.006 & 0.91 & \texttt{\_Cur, \_cur, Cur, \_CUR, cur, CUR, \_Kur}\\
 \bottomrule[0.1em]
\end{tabularx}
}
\captionof{table}{\textbf{Examples of Alphabet Latents.} Latents from layer 25 of Gemma 2 2B that promote or suppress tokens sharing an initial letter. ``Metric'' is the fraction of top 100 affected tokens starting with that letter.} 
\label{table:alphabet}
\vspace{-2mm}
\end{minipage}
\end{figure}

We find that some nullspace latents produce much larger entropy shifts than the random baseline, indicating that they encode signals relevant to entropy modulation. In particular, latent 14325 has a disproportionate impact on output entropy. To test whether this signal is used by the model in conjunction with RMSNorm scaling (as in \citet{stolfo2024confidence}), we repeat the ablation while freezing the RMSNorm scaling coefficient. \cref{fig:nullspace_latents}c shows that the entropy change diminishes under this intervention, suggesting that the model uses this direction to modulate entropy via RMSNorm. Furthermore, \cref{fig:entropy_correlation} shows that the combined activation of the antipodal pair formed by latents 13748 and 14325 is strongly correlated with output entropy, further supporting this interpretation.

While these results highlight the functional role of specific nullspace latents in entropy regulation, not all latents in this subspace behave similarly. Some exhibit negligible impact on entropy when ablated. We speculate that these may track different internal signals--one such candidate is the attention sink signal, which has also been associated with the $\WU$ nullspace \cite{cancedda-2024-spectral}. Overall, these experiments provide mechanistic evidence that nullspace latents correspond to internal model computations.

\subsection{Alphabet Latents}
\label{sec:alphabet}
We identify a class of dense latents that selectively boost or suppress large sets of tokens sharing the same initial letter. Unlike prior work that linked latents to the \emph{current} token’s first letter \cite{chanin2024absorptionstudyingfeaturesplitting}, these instead relate to the \emph{next} token’s initial character.

To discover these latents systematically, we examine each latent’s top 100 positive and negative logit contributions by projecting its decoder weights onto the vocabulary space. Then, we collect the corresponding tokens, and select latents where either set contains at least 90\% of tokens starting with the same character (excluding the space character ``\texttt{\_}''). At layer 25, this procedure yields 114 such latents, of which 21 have activation density >0.1, accounting for 20\% of all dense latents. These latents span a range of antipodality scores and activation densities, but notably appear as high-density features only at the model’s final layer. We provide some examples from this layer in \cref{table:alphabet}.

Interestingly, we observe multiple latents for each letter, varying in specificity: some target a broad set of short tokens sharing only the first letter (e.g., ``b'' or ``c''), while others focus on longer tokens sharing a multi‐letter prefix (e.g., ``br'' or ``cu''). We attribute this granularity to feature splitting \cite{bricken2023monosemanticity} possibly driven by n‑gram frequency, which yields latents with differing activation densities.
These latents illustrate how SAEs dedicate dense units to encode output-specific signals related to next-token lexical structure.




\subsection{Meaningful-Word Latents}
\label{sec:noun_latents}
\begin{figure}
    \centering
    \includegraphics[width=1\linewidth]{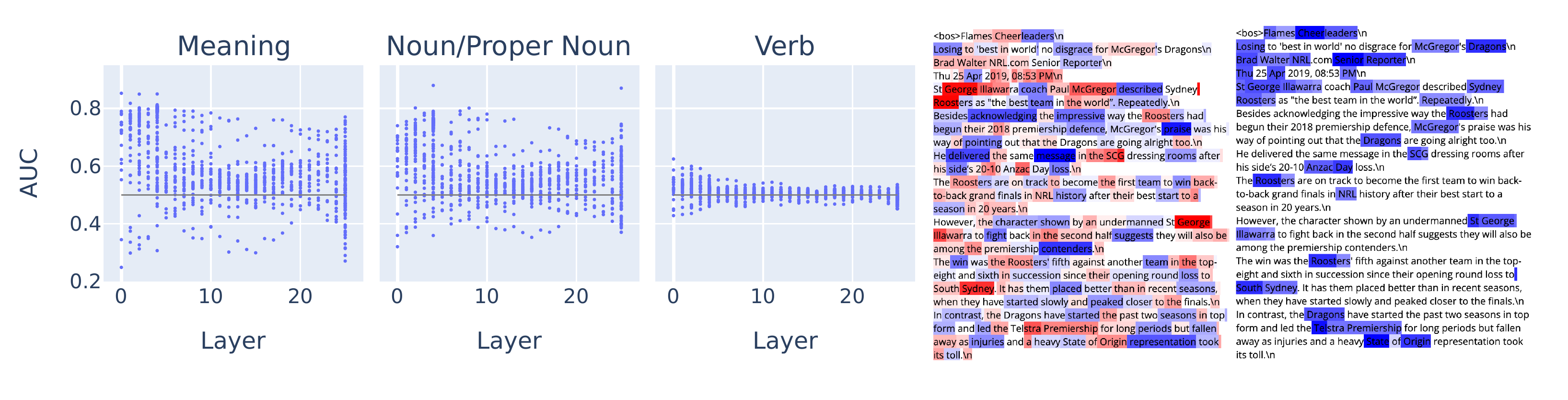}
    \caption{\textbf{Meaningful-Word Latents.} (Left) AUCs of predicting feature firing, from whether the POS tag is within the specific category. ``Meaningful word'' and ``noun/propernoun'' are good predictors, while other categories like ``verb'' are less predictive. (Middle) Example of L2: pair 15089 (blue), 13092 (red) firing patterns on a document, where 15089 fires on ``meaning-heavy'' words while 13092 fires on proper nouns and functional words (the, in, a). (Right) Example of L3: 7507 firing patterns, where it fires selectively on proper nouns.}
    \label{fig:pos}
\end{figure}
The next class of latents that we investigate are those whose firing can be well predicted by the part-of-speech (POS) tag of the token. We create a reduced set of high-level tags from the Brown Corpus \citep{francis1979brown} by combining similar tags (e.g., combining plural and singular forms of nouns),\footnote{See \cref{tab:ptb_map} in \cref{app:meaningful} for our full mapping.} and capture dense latent activations on 10k sentences ($\approx$ 200k tokens) from the corpus. Then, for each latent, we calculate the AUC-ROC of predicting the binary latent activations given the binary vector of whether a token is within the high-level POS category. Intuitively, this AUC reflects how well the interpretable linguistic category \textit{predicts} the latent.

We find that even these high-level groupings are not enough to achieve a high AUC (\cref{fig:pos,fig:additional_pos}), and propose a further grouping of these tags into ``meaningful words'', where a token is considered a ``meaningful word'' if it is one of $\{$nouns, proper nouns, verbs, adjectives, adverbs$\}$. The resulting binary-binary predictor has a decent AUC (\cref{fig:pos}) of $\approx0.8$ for many dense latents in early layers, suggesting that the model contains a dense subspace tracking the presence of these meaningful words. 


\subsection{PCA Latents}
\label{sec:pca_latents}
Since the top principal components (PCs) are a large fraction of the variance of the activations, one might expect an SAE to learn dense latents that simply reconstruct this subspace. However, we find that this hypothesis is only partly the case: as shown in \cref{fig:pc15}, an antipodal pair of latents consistently reconstruct most of the first PC (cosine similarity $> 0.75$), but other latents do \textit{not} have a large norm percentage in the top PC, even up to the top 5 PC components. The top PC-aligned latents are generally not immediately interpretable and do not fall into any of our classes above. Interestingly, decreasing or increasing the SAE $L_0$ and dictionary size does not eliminate PC-aligned latents nor result in significantly more of them (\cref{fig:pcdiff}).\looseness=-1

\subsection{Layer-wise Dynamics}

\begin{figure}[t]
    \centering
    \includegraphics[width=0.4\textwidth]{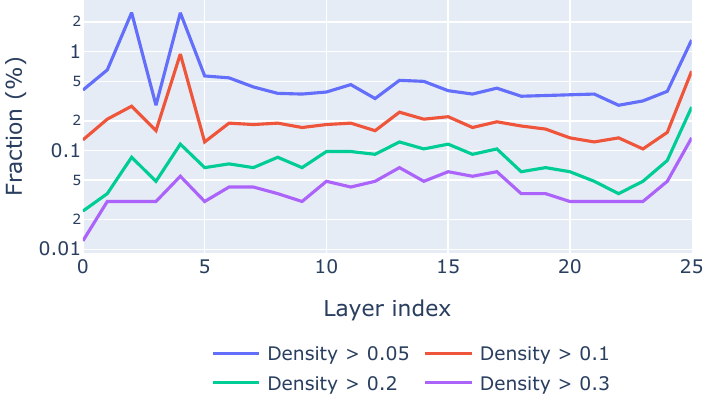}
    \includegraphics[width=0.4\textwidth]{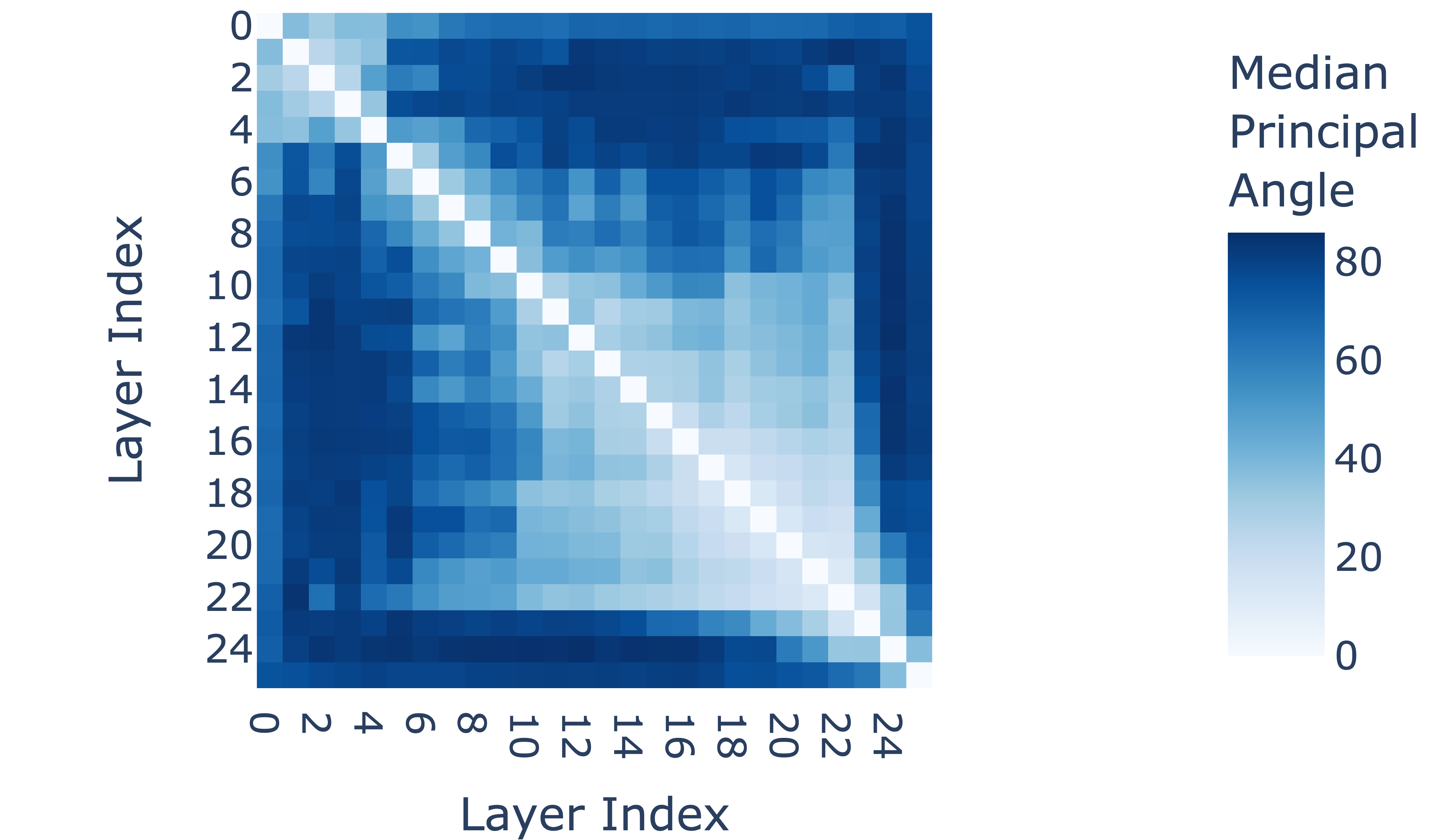} 
\caption{\textbf{Layer-wise Dynamics of Dense Latents.} 
(a) Fraction of dense latents (at various density thresholds) across residual stream SAEs at different layers of Gemma 2 2B.  
(b) Median principal angles between dense-latent subspaces, showing a shift in subspace structure from early to late layers.}
  \label{fig:across_layers}
\end{figure}

As noted in the taxonomy of dense latents above, and visualized in \cref{fig:master_taxonomy}, each class of dense latents is found in specific layer ranges. Dense latents in early layers have more token-dependent activations and track positional information, those in middle layers represent more conceptual directions, and those in the final layers are mostly mechanisms that the model uses to control its output. Inspired by these observations, in this section we further examine layer-wise characteristics of dense latents.

\paragraph{Number of Dense Latents.} First, we study how the number of dense latents changes across different layers of the model. \cref{fig:across_layers}a illustrates the fraction of latents exceeding density thresholds of 0.05, 0.1, 0.2, and 0.3 at each layer. In the early layers (0-4), we observe transient spikes in latents just above the 0.05 and 0.1 thresholds. These latents are largely the part-of-speech related latents in \cref{sec:noun_latents}. The absence of similar spikes at the 0.2 and 0.3 thresholds suggest that these early fluctuations arise from SAE training variability rather than fundamental differences in the information encoded at different points of the model’s residual stream. Across the middle layers (5–23), the fraction of dense latents is remarkably stable for all thresholds. Finally, the model’s last two layers exhibit an increase in the number of dense latents, indicating a final emergence of dense features prior to unembedding.


\paragraph{Consistency of the Dense Subspace.}
We next ask whether the subspace spanned by dense latents remains stable across layers or varies over the model. For each pair of layers, we compute the principal angles between the subspaces defined by latents with density $>0.2$, then take the median angle as a summary statistic: values near 0° indicate largely overlapping subspaces, while values near 90° indicate dissimilarity. \cref{fig:across_layers}c visualizes these median angles for every layer pair of Gemma 2 2B.\footnote{We find that using a slightly higher density threshold ($0.2$) makes the subspace similarity pattern more pronounced. The same plot with a lower threshold ($0.1$) is shown in \cref{app:angles}, showing the same clustering trend but with reduced overall similarity.}
Three clusters emerge. Layers 0-4 share a common dense subspace (low angles). This shifts in the middle of the model (layers 10–22), where a new stable subspace persists (mutually low angles). Finally, the last few layers exhibit a pronounced change (large angles relative to earlier layers), consistent with the rise of alphabet and nullspace latents before the unembedding.



\section{Related Work}
\paragraph{Sparse Autoencoders.} Transformer models are thought to represent features as linear directions in activation space \cite{mikolov, nips2016_a486cd07, elhage2021mathematical, nanda-etal-2023-emergent, pmlr-v235-park24c, olah2024lrh}, with many more features than neurons, leading to \emph{superposition} \cite{olah2020zoom,elhage2022superposition}. Early work explored sparse dictionary learning to interpret these representations \cite{OLSHAUSEN19973311, faruqui-etal-2015-sparse, arora-etal-2018-linear, zhang2021wordembeddingvisualizationdictionary}. More recently, sparse autoencoders (SAEs; \citealp{ng2011sparse}) have emerged as a scalable and effective implementation of sparse dictionary learning for transformer-based models \cite{yun-etal-2021-transformer, bricken2023monosemanticity, huben2024sparse, rajamanoharan2024improving, rajamanoharan2024jumpingaheadimprovingreconstruction, kissane2024interpreting, bussmann2025learningmultilevelfeaturesmatryoshka} that can recover meaningful and causally important features \cite{templeton2024scaling, gao2025scaling, marks2025sparse}.

\paragraph{Interpreting SAE Latents.}
As SAEs have gained traction, recent work has focused on interpreting the meaning of their latent features \cite{chanin2024absorptionstudyingfeaturesplitting, leask2025sparse}. Building on the neuron interpretation methodology in \cite{bills2023language}, several recent works interpret SAE latents systematically. 
\citet{templeton2024scaling} propose a rubric-based evaluation method in which a language model (Claude 3 Opus) scores how well a proposed feature description aligns with the contexts on which the latent activates. Similarly, \citet{paulo2024automaticallyinterpretingmillionsfeatures} propose a pipeline in which natural language interpretations for SAE latents are matched with different contexts and used by an LLM in different tasks that evaluate how good the interpretations are in predicting activating and non-activating contexts. Other recent efforts explore automated interpretation approaches based on self-interpretation strategies \cite{self-explaining}.
A recurring observation across multiple studies is \emph{dense} latents, which activate on more than 10\% or even 50\% of tokens \cite{topkvsgated, rajamanoharan2024jumpingaheadimprovingreconstruction}.
\citet{anthropic-interpretable-dense-features} take the 10 most densely activating latents in a cross-layer Transcoder trained on Claude and attempt to manually interpret them, finding plausible interpretations (e.g., ``activates on commas,'' ``activates on non-terminal tokens in multi-token words'') for 6 of the 10 features. In contrast, \citet{removing-dense-latents} view dense latents as an undesired phenomenon and propose a frequency-based regularizer to discourage their emergence during training.
Whether these latents reflect meaningful internal computations or arise as undesirable artifacts was up until our work an open question. 

\paragraph{Dense Language Model Representations.}
Prior work has also identified dense signals in language model representations more broadly (i.e., components that encode information consistently across many tokens). \citet{gurnee2024universalneuronsgpt2language} present a taxonomy of universal neurons that appear across GPT-2 models trained with different seeds. Among these, they identify neurons that encode positional information. \citet{chughtai2024understanding} similarly identify dense positional features in an SAE trained on GPT-2’s layer 0, though they do not explicitly analyze their activation density. Finally, \citet{stolfo2024confidence} describe neurons that regulate model confidence by tracking entropy and connect them to a component of the residual stream aligned with the quasi-nullspace of the unembedding matrix.


\section{Discussion, Limitations \& Conclusion}
\label{sec:conclusion}

Our work shows that dense SAE latents discover intrinsically dense features in the underlying language model representations. This challenges recent efforts that aim to remove dense latents with ad-hoc penalties in the SAE loss function \cite{removing-dense-latents}. Our results motivate future feature-extraction mechanisms that are able to find features that are not necessarily sparse. For example, such techniques might include SAE designs that allocate autoencoder capacity for representing dense subspaces, approaches that optimize circuit sparsity, or techniques like APD \cite{braun2025interpretability} that focus on parameter sparsity.




\textbf{Limitations.} Although our work identifies some classes of dense latents, we do not claim that all dense latents encode interpretable or meaningful signals. We hypothesize that some dense latents are a noisy aggregation of sparse features rather than a ``true'' dense feature, and distinguishing between these remains an open challenge. Moreover, dense latents may learn a basis that spans \textit{but does not align with} the set of true dense model representations, since dense latents co-occur extremely frequently, and a linear combination of the ``true'' basis works for reconstruction too.

Despite consistently observing the antipodality trend across both TopK and JumpReLU SAEs and across models (Gemma 2 2B and GPT-2 Small), our interpretability analysis primarily focuses on JumpReLU SAEs trained on Gemma 2 2B, using a single dictionary size and sparsity constraint per layer. Future work could broaden analysis to more models, SAE architectures, and SAE sparsities.

Most notably, we have explained less than half of dense SAE features. We view understanding the rest of these latents as exciting future work that could provide insight into frequently-active, fundamental mechanisms and representations in language models.

\section*{Acknowledgments}
We would like to express our gratitude to Arthur Conmy, Neel Nanda, and Vilém Zouhar for their valuable feedback and insightful discussions at different points during the development of this project. AS acknowledges the support of armasuisse Science and Technology through a CYD Doctoral Fellowship.

\bibliography{bibliography}
\bibliographystyle{plainnat}


\newpage
\appendix

\begin{figure}[t]
    \centering
    \includegraphics[width=0.325\textwidth]{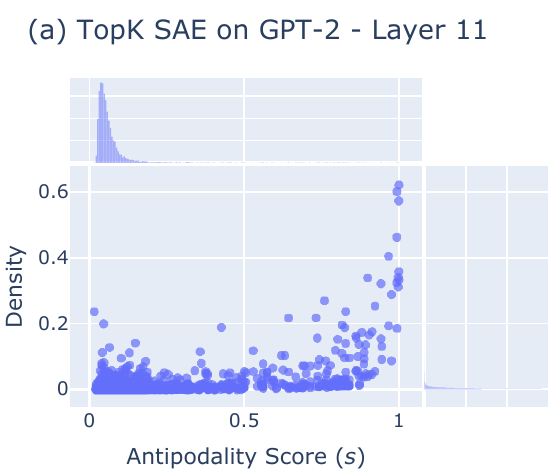}
    \includegraphics[width=0.325\textwidth]{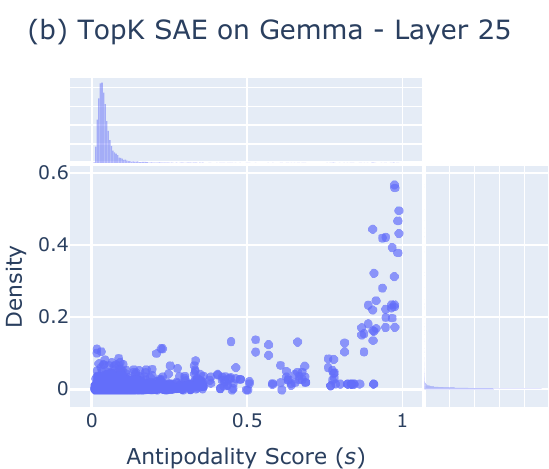}
    \includegraphics[width=0.325\textwidth]{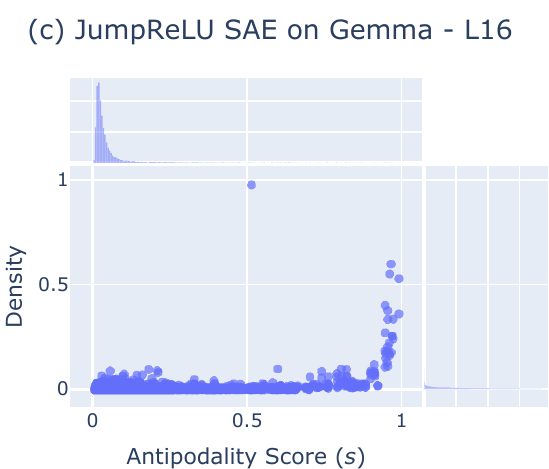}
    
\caption{\textbf{Additional Antipodality Plots.} Antipodality scores vs.\ activation density for (a) TopK SAE on GPT-2 (Layer 11), (b) TopK SAE on Gemma 2 2B (Layer 25), and (c) JumpReLU SAE on Gemma 2 2B (Layer 16). Across all configurations, dense latents tend to have high antipodality scores.
}
  \label{fig:antipodality_appendix}
\end{figure}

\begin{figure}[t]
    \centering
    \includegraphics[width=0.325\textwidth]{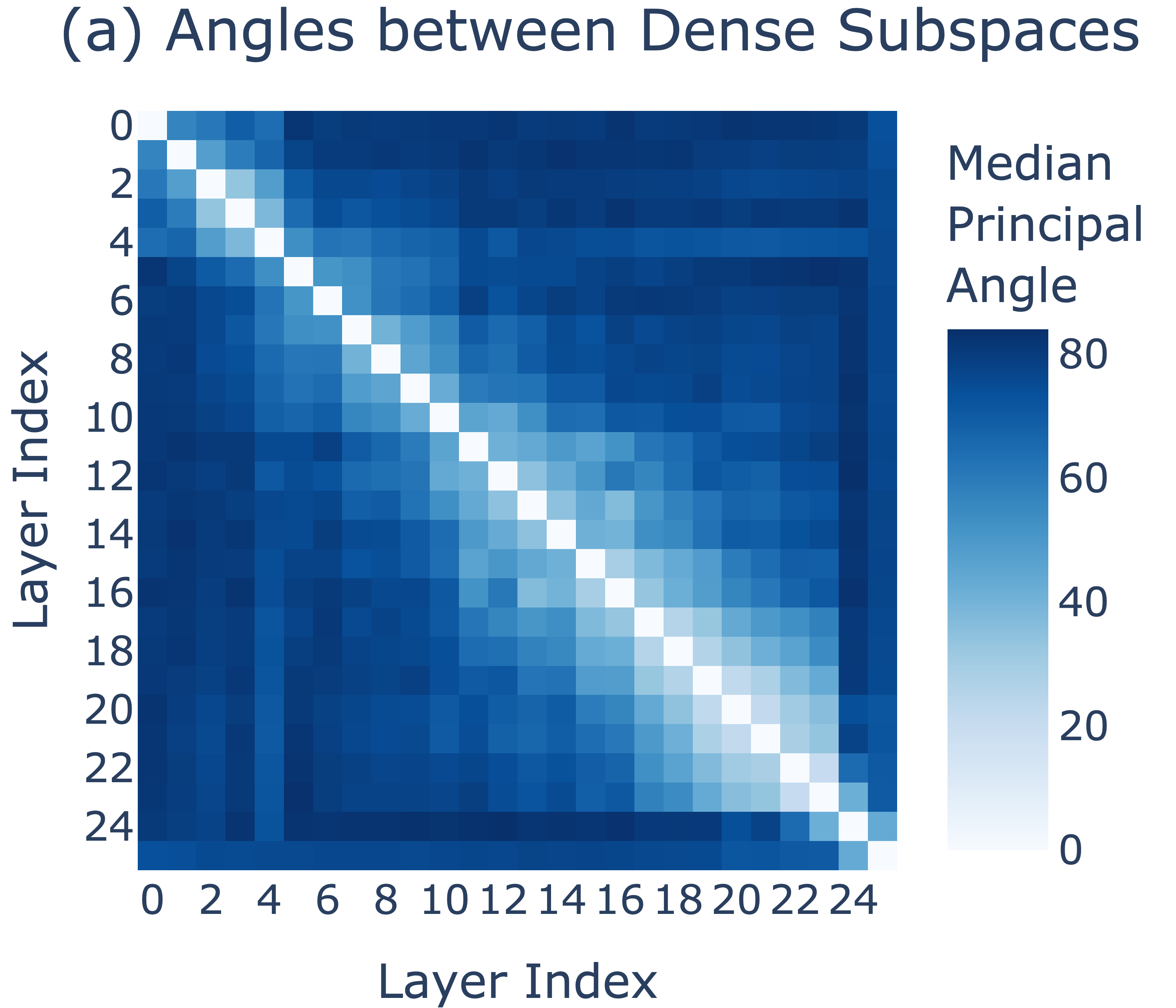}
    \includegraphics[width=0.325\textwidth]{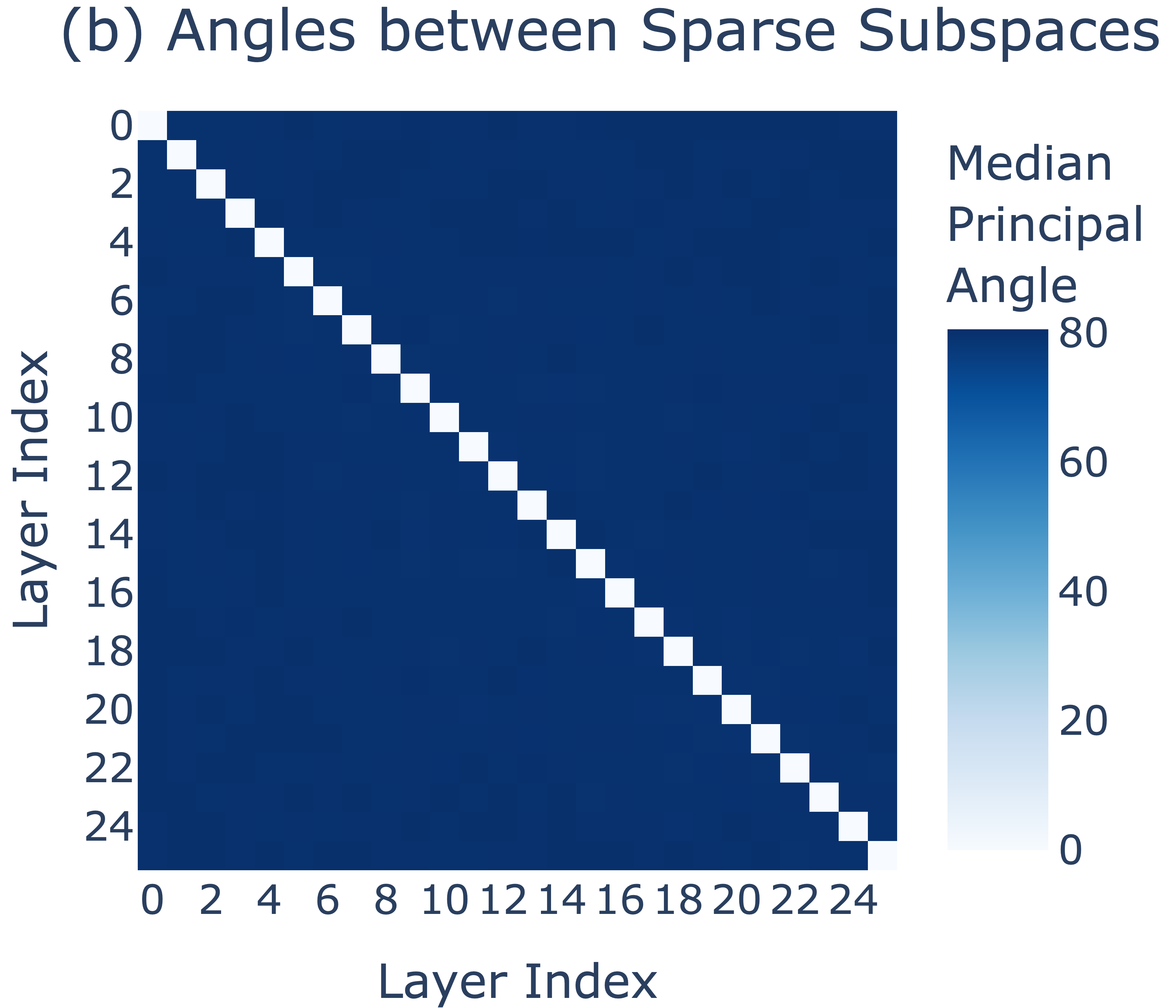}
    \includegraphics[width=0.325\textwidth]{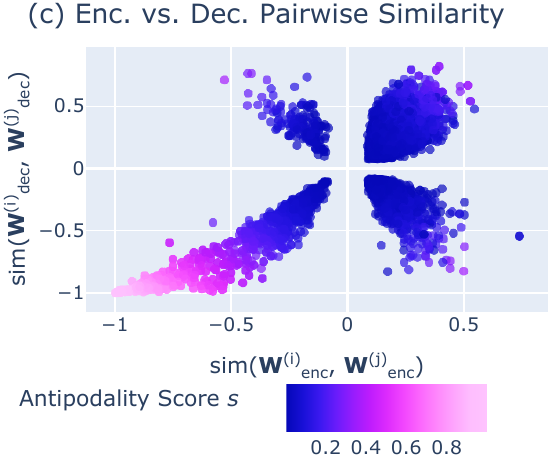}
\caption{\textbf{Additional Analyses.} (a) Median principal angles between dense-latent subspaces (density>0.1) across layers. (b) Principal angles between randomly selected non-dense latent subspaces. (c) High antipodality score occurs when encoder and decoder weights are nearly opposite.
}
  \label{fig:additional_res}
\end{figure}

\section{Additional Results}
\subsection{Antipodal Pairing in Different SAEs}
\label{app:antipodality}
\cref{fig:antipodality_appendix}, we report antipodality scores (computed as in \cref{eq:antipodality_score}) for dense latents in three additional SAEs: two TopK SAEs that we trained on the residual streams of GPT-2 (layer 11) and Gemma 2 2B (layer 25), and a JumpReLU SAE from the Gemma Scope suite trained on an earlier layer (16). In all cases, we observe the same trend highlighted in \cref{fig:general_characteristics}c: high-density latents cluster at high antipodality scores, forming near-antipodal pairs that reconstruct specific directions in residual space.

\begin{wrapfigure}{r}{0.325\textwidth}
    \centering
    \vspace{-\intextsep}
    \includegraphics[width=\linewidth]{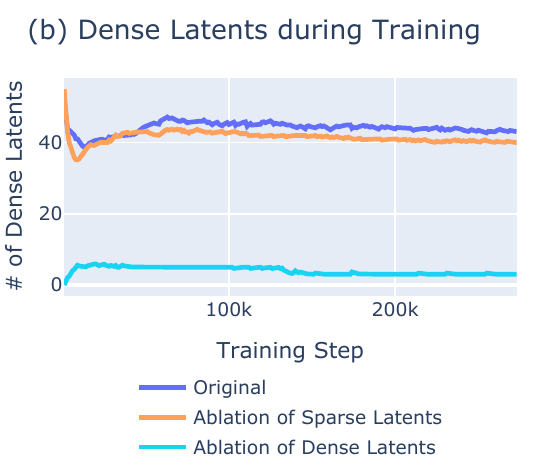}
    \caption{\textbf{Dense Latents During Training.} Dense latent counts stabilize early in training.}
    \label{fig:training}
    \vspace{-\intextsep}
\end{wrapfigure}

\subsection{Dense Latents During Training}
\label{app:training}
In \cref{fig:training}, we visualize the number of dense latents (activation frequency $>0.1$) over training steps for each SAE configuration in our ablation experiment described in \cref{sec:dense_ablation}. All curves converge within the first $\sim$100k steps and remain stable throughout training. This early plateau suggests that dense latents are not a product of late-stage optimization noise, but rather emerge early and persist, indicating that they reflect consistent structure in the residual stream rather than transient artifacts.

\subsection{Angles Between Residual Stream Subspaces}
\label{app:angles}
In \cref{fig:additional_res}, we provide further analysis of the evolution of dense latent subspaces across layers. Panel (a) shows the median principal angle between the subspaces spanned by latents with density $>0.1$ at each pair of layers in Gemma 2 2B. These results follow the trend observed in \cref{fig:across_layers}c (based on a $>0.2$ cutoff), revealing distinct subspace clusters in the early, middle, and late layers. However, the overall similarity between subspaces is lower here, reflecting the greater variability introduced by including moderately dense latents (density 0.1-0.2).

For comparison, panel (b) reports the same metric computed on subspaces spanned by 100 randomly selected non-dense latents per layer. As expected, these subspaces exhibit minimal overlap, with median principal angles near 90° across all layer pairs, confirming that the structure observed in the dense-latent subspaces is nontrivial.

\subsection{Pairwise Similarity Between Latents' Weights}
\label{app:cosine_sim}
In \Cref{fig:additional_res}c, we report for each latent $i$, the maximum-magnitude cosine similarity of its encoder and decoder weights with any other latent $j$.
In particular, we show $\operatorname{sim}(\Wenci, \Wencj)$ and $\operatorname{sim}(\Wdeci, \Wdec^{(k)})$, where $j = \argmax_{l\neq i} (|\operatorname{sim}(\Wenci, \Wenc^{(l)})|)$ and $k= \argmax_{l\neq i} (|\operatorname{sim}(\Wdeci, \Wdec^{(l)})|)$.
We find that the antipodality score $s$ approaches 1 only when both encoder and decoder similarities are close to $-1$.

\subsection{Similarity with SAE Bias}
\label{app:bias}
\begin{figure}[t]
    \centering
    \includegraphics[width=0.6\linewidth]{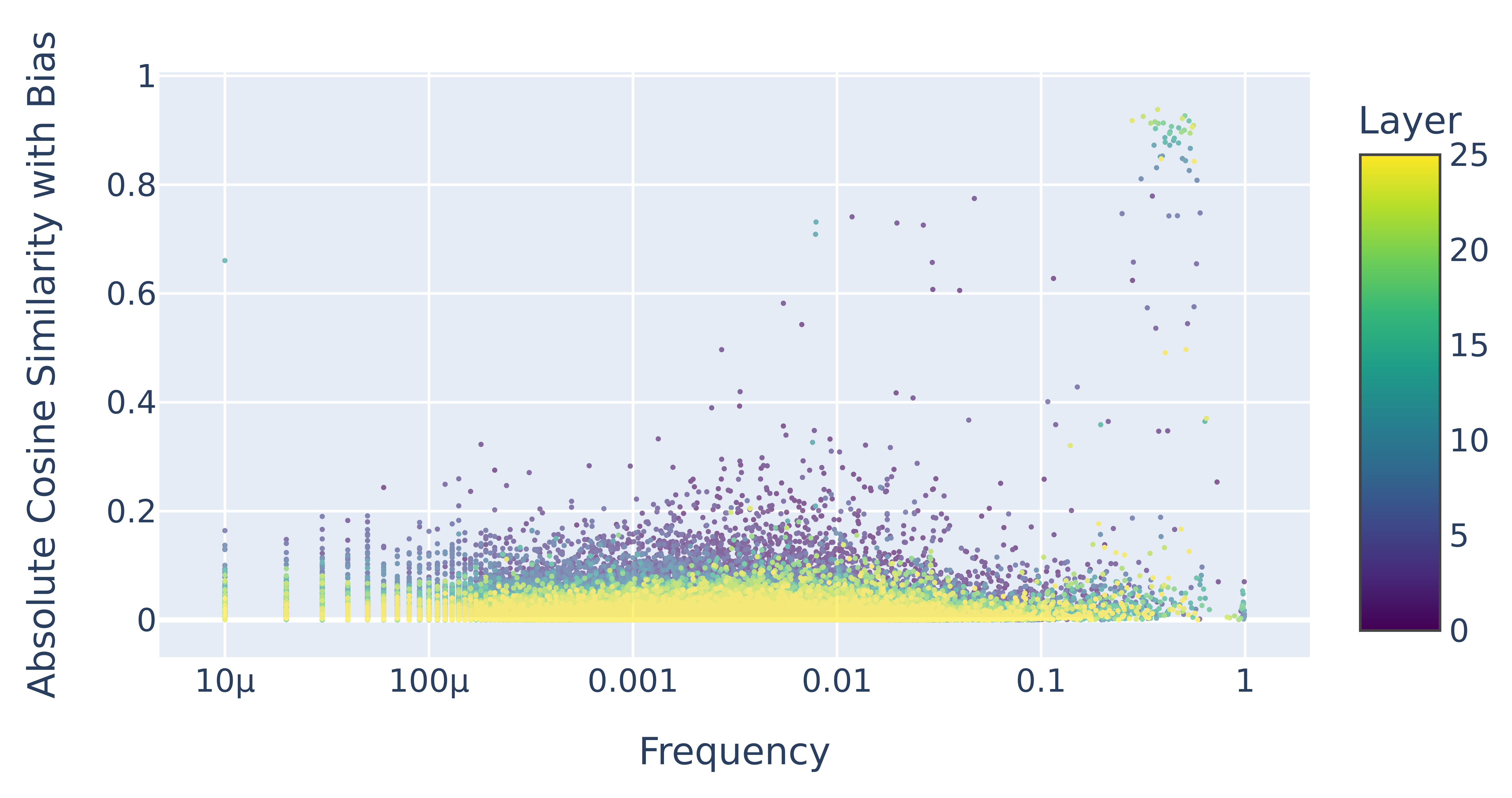}
    \caption{Plot of absolute cosine similarity of all SAE decoder vectors at all layers with that layer's decoder bias. We observe a group of dense latents in the upper right corner that have high frequency and align with the bias.}
    \label{fig:bias}
\end{figure}

To investigate the relationship between dense latents and the SAEs’ bias terms, we compute the cosine similarity between each SAE decoder vector and the corresponding layer’s decoder bias. Figure \ref{fig:bias} shows the absolute cosine similarity for all latents across layers as a function of activation frequency. We observe a small but distinct group of dense latents (upper-right region) that strongly align with the bias.

\begin{figure}[t]
    \centering
    \includegraphics[width=0.325\textwidth]{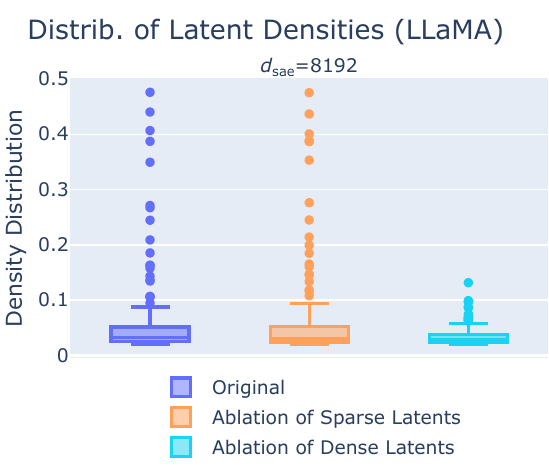}
    \includegraphics[width=0.325\textwidth]{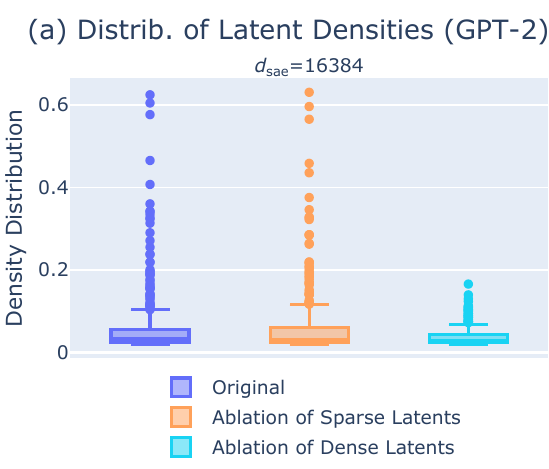}
\caption{\textbf{Dense-subspace ablations on LLaMA-3.2-1B and GPT-2 Small.} For each model’s final layer we train a baseline SAE (blue), retrain after ablating the subspace spanned by dense latents (teal), and retrain after ablating an equally sized subspace of non-dense latents (orange). Only removing the dense-latent subspace collapses the high-density tail. 
}
\label{fig:additional_ablations}
\end{figure}

\subsection{Additional Dense-latent Subspace Ablations}
\label{app:dense_ablations}

\cref{fig:additional_ablations} replicates the dense-subspace ablation from \cref{sec:dense_ablation} on two additional models: LLaMA-3.2-1B \citep{grattafiori2024llama3herdmodels} and GPT-2 Small \citep{radford2019language}. For each model we (i) train a baseline SAE, (ii) retrain after zero-ablating the subspace spanned by dense latents, and (iii) retrain after ablating an equally sized subspace of randomly chosen non-dense latents. All experiments are run at the final residual-stream layer. In both cases, removing the dense-latent subspace collapses the high-density tail---yielding almost no dense latents---whereas ablating a sparse subspace leaves the distribution essentially unchanged. These replications mirror the Gemma 2 2B result and further support that dense latents reflect an intrinsic residual-stream subspace rather than a training artifact.

\section{Experimental Details}
\label{app:exp_details}
For the experiment in \cref{sec:dense_ablation}, we trained TopK SAEs \cite{gao2025scaling} on the residual stream activations at layer 25 of Gemma 2 2B using 1 billion tokens from the OpenWebText corpus \cite{gokaslan2019openweb}. Training followed the default configuration of the \texttt{Sparsify} library,\footnote{\url{https://github.com/EleutherAI/sparsify}} and experiment tracking was conducted using Weights \& Biases.\footnote{\url{https://wandb.ai}}
The ablation experiment on nullspace latents described in \cref{sec:nullspace} was performed on a 10k-token subset of the C4 corpus \cite{c4}. Analyses throughout the paper were conducted using the Gemma Scope SAEs \cite{lieberum2024gemmascopeopensparse} with 16k latents  trained on the residual stream of Gemma 2 2B.
All experiments were implemented in \texttt{PyTorch} \citep{NEURIPS2019_bdbca288}, with model inspection tools from the \texttt{TransformerLens} library \citep{nanda2022transformerlens}. Data processing used \texttt{NumPy} \citep{harris2020array} and \texttt{Pandas} \citep{mckinney-proc-scipy-2010}, and figures were generated with \texttt{Plotly} \citep{plotly}.

\section{Compute Resources Used}
\label{app:compute_used}
We expect the experiments for training SAEs, capturing SAE activations and generating completions with Gemma 2 2B to be able to be run in about 30 A6000 hours. The LLM judging experiments take less than USD \$20 through OpenRouter with Gemini 2.5 Flash Preview \citep{google2025gemini}.

\section{Broader Impact}
\label{app:broader_impact}
Our work focuses on interpreting language models, an important component of building safer and more reliable systems. SAEs in particular are a popular technique for understanding language models, and through investigating dense latents, we can both better inform SAE design, and better understand language model internals.

We do not foresee any negative impacts of our work.

\begin{figure}[t]
    \centering
    \includegraphics[width=\linewidth]{figures/proximity.pdf}
    \caption{\textbf{Position Latents}. We identify position latents by computing their Spearman correlation $\rho$ with relevant text boundaries. We classify a latent as belonging to a certain category when $|\rho|>0.4$.}
    \label{fig:position}
\end{figure}

\begin{figure}[t]
    \centering
    \includegraphics[width=1\linewidth]{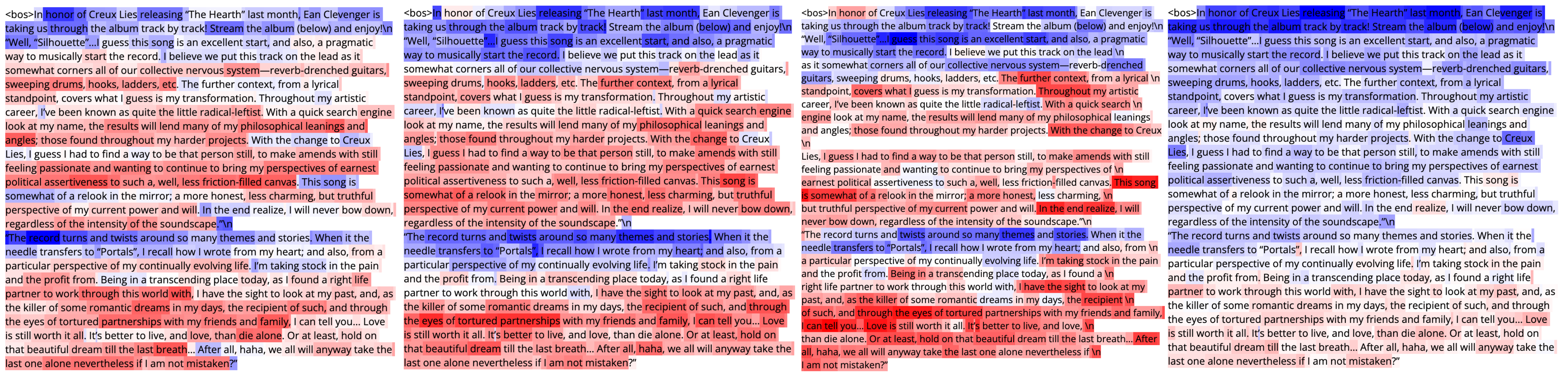}
    \caption{Examples of position latents in layer 5. Deep blue represents positive projection along decoder vector, and deep red represents negative. (1) L5:4341 is a sentence-tracking latent, that lights up consistently on beginnings of sentences. It has strong activations for topic sentences too. (2) L5:8680 is a paragraph-tracking latent, that lights up on beginnings of paragraphs. (3) L5:8680 is agnostic to artificially adding formatting newlines, showing it is encoding true paragraph position. (4) L5:697 is a context-position-tracking latent.}
    \label{fig:position-examples}
\end{figure}
\begin{figure}[t]
    \centering
    \includegraphics[width=0.6\linewidth]{figures/newline_period.pdf}
    \caption{Spearman correlation for period against Spearman correlation for newline.}
    \label{fig:newline_period}
\end{figure}

\section{Additional Taxonomy Results}

\subsection{Classification of dense latents}
\label{app:class}
In our taxonomy, we identify dense latents using automated tests. We do not expect these tests to be perfect for a variety of reasons---for instance, dense latents not lining up perfectly with the ``true'' feature basis due to learning a linear combination basis, and the fundamental difficulty of designing true, causal tests. However, for the purposes of illustration, we choose reasonable cutoffs for each test to create \cref{fig:master_taxonomy}, listed below.

\begin{itemize}
    \item Position latents: Spearman correlation of $|\rho|>0.4$ for the relevant text boundary.
    \item Context-binding latents: Fraction of successful flips $> 0.75$.
    \item Nullspace latents: $>0.2$ of encoder weight in bottom 10 $\mathbf{W_U}$ singular vector subspace.
    \item Alphabet latents: Top 100 or bottom 100 logit contributions contain at least 90\% of tokens starting with same character.
    \item Meaningful-word latents: AUC of using ``is meaningful word'' to predict ``feature fires'' $>0.75$.
    \item PC-aligned latents: cosine similarity with top PC $>0.75$.
\end{itemize}

Very few dense latents (3.6\% across layers) fall in >1 category based on our automated tests to find them, with the most common clashes being between sentence- and paragraph- tracking (see \cref{app:position}), and between several categories and meaningful-word latent. For the purposes of illustration, we break ties according to the priority (from highest to lowest): \{context-tracking, sentence-tracking, alphabet, nullspace, context-binding, paragraph-tracking, meaning, PCA\} based on our confidence in our automated tests.

\subsection{Position latents}
\label{app:position}

The observation in \cref{fig:newline_period} that period-tracking and newline-tracking latents are hard to distinguish also relates to our discussion in \cref{sec:conclusion} that because the sparsity incentive is low for these dense latents, they may not be perfectly aligned to ``true'' model dense features, and may instead be a linear combination of two related features.

\begin{figure}[t]
    \centering
    \includegraphics[width=1\linewidth]{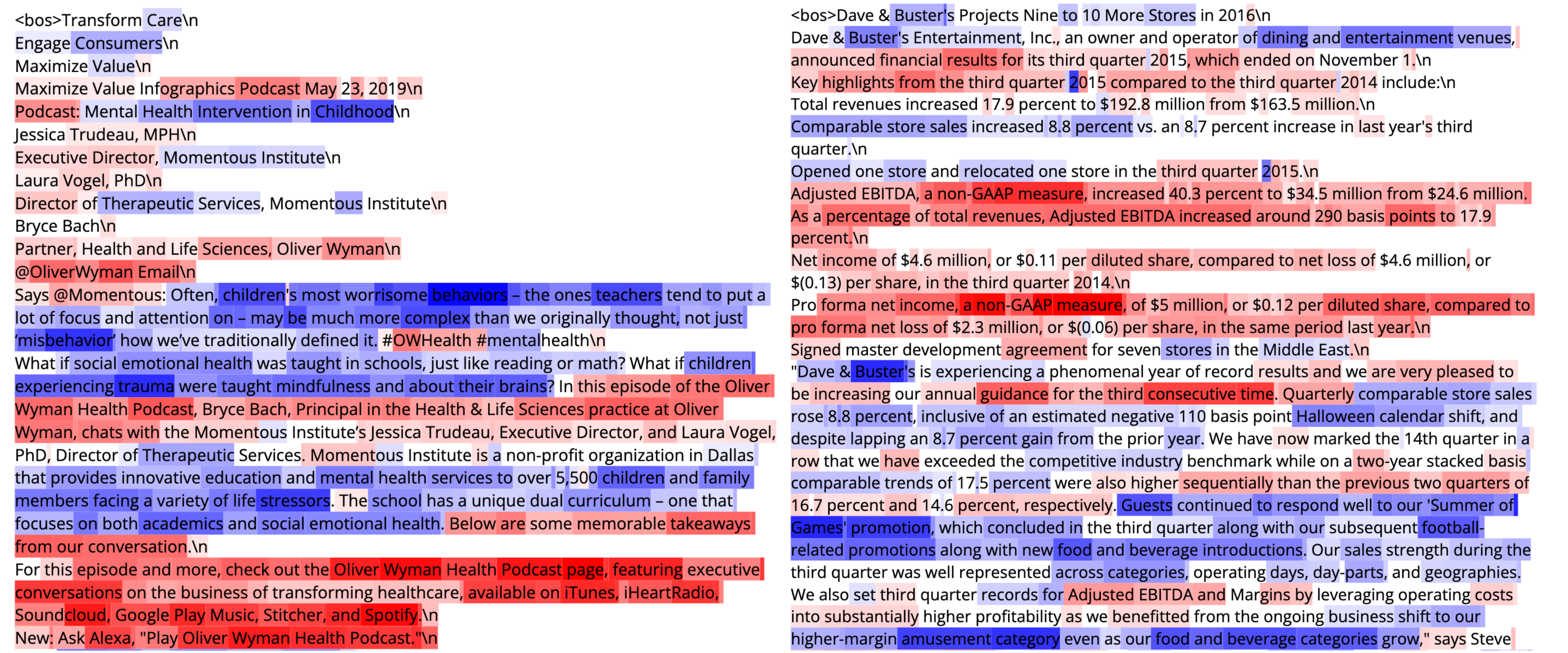}
    \caption{L13: 15275 (blue) and 11449 (red), which has 81.5\% correct flips. In these two examples, 15275 fires on children's mental health (left) and Dave \& Buster's promotions (right), while 11449 fires on mentions of the podcast (left) and financial measures (right).}
    \label{fig:13-15275-11449}
\end{figure}

\begin{figure}[t]
    \centering
    \includegraphics[width=1\linewidth]{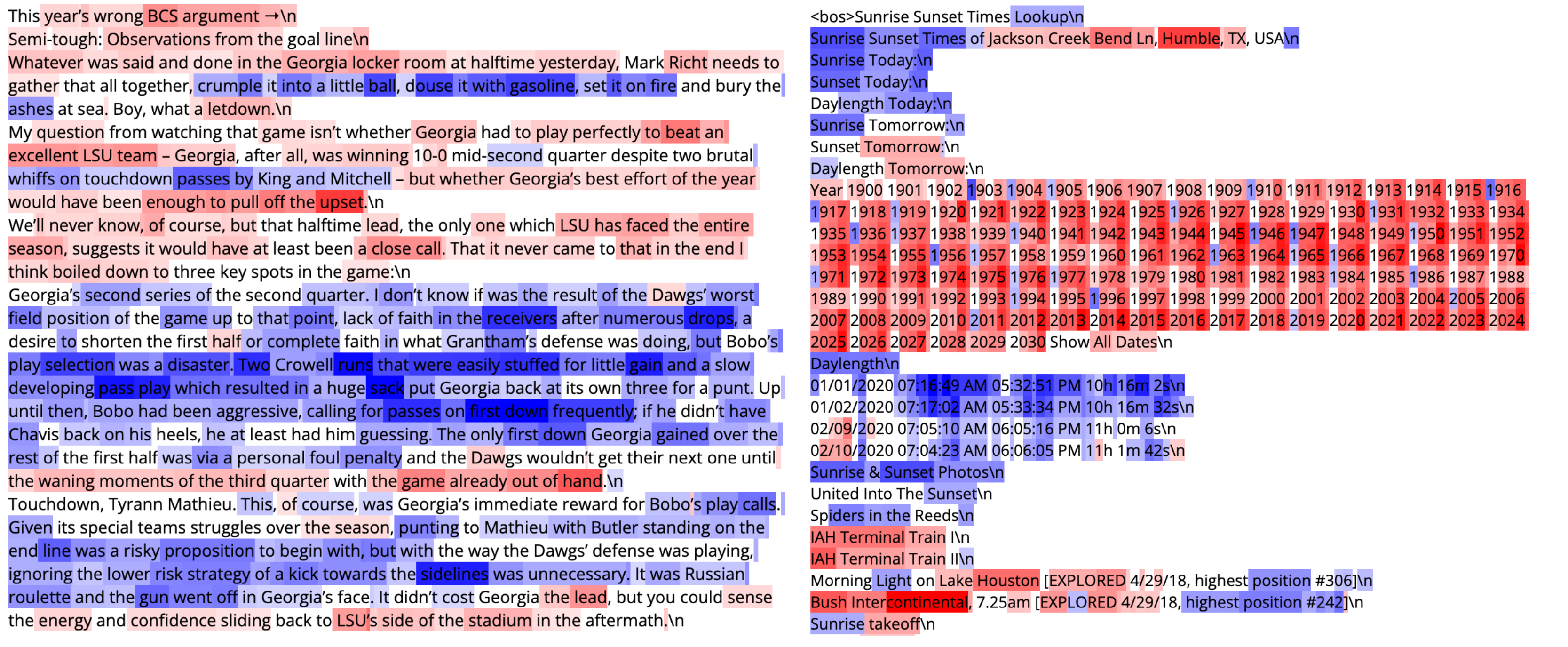}
    \caption{L12: 14906 (blue) and 14599 (red), which has 76.5\% correct flips. In these two examples, 14906 fires on descriptions of the game (left), and text or numbers related to sunrise (right), while 14599 fires on the teams and winning/losing (left), and years or locations (right).}
    \label{fig:ll}
\end{figure}

\subsection{Interpreting context-binding latents}
\label{app:autointerp}

We attempt to interpret mid-layer latents that exhibit coherent chunk-level activations in two ways:
\begin{enumerate}
    \item \textbf{All-context}: Following existing autointerp methods \citep{paulo2024automaticallyinterpretingmillionsfeatures}, we sample 10 activating and 10 non-activating phrases from an entire corpus and ask an LLM (Gemini 2.5 Flash) to generate an explanation. We repeat this 100 times to generate 100 explanations.
    \item \textbf{In-context}: We instead sample 10 activating and 10 non-activating phrases from the \textit{same context}. We repeat this 100 times (using 100 different contexts) to generate 100 explanations.
\end{enumerate}
When examples are drawn from the same context, the explanations are specific but highly diverse across contexts. When examples are drawn from different contexts, the explanations become vague or generic (\cref{tab:contextbinding_dense_combined}). This drop in specificity across contexts is somewhat expected, since explanations for any latent may overfit the context. However, doing the same for sparse latents (\cref{tab:contextbinding_sparse}), we see that a ``good'' sparse latent would have similar explanations with both in-context and all-context examples, aligning with the usual assumption that SAEs learn directions that represent a concept in the model.

It is difficult to rule out the possibility that these dense latents represent an uninterpretable abstract feature the model learns. However, the steering experiment seems to cause the relevant specific concepts to be brought up during generation, supporting the ``binding'' hypothesis that there are dense directions that do not represent a fixed concept but rather are used in the model's computation.

\subsection{Additional examples of context-binding latents}
\label{app:context_binding_examples}
We include two additional examples of context-binding latent pairs with high flip score: layer 13 pair (15275, 11449) (\cref{fig:13-15275-11449}) and layer 12 pair (14906, 14599) (\cref{fig:ll}). For each pair, we show two example contexts where they are active, illustrating how each latent activates on specific but context-dependent concepts, and that latents in a pair do not co-activate.

\begin{figure}[t]
    \centering
    \includegraphics[width=0.24\textwidth]{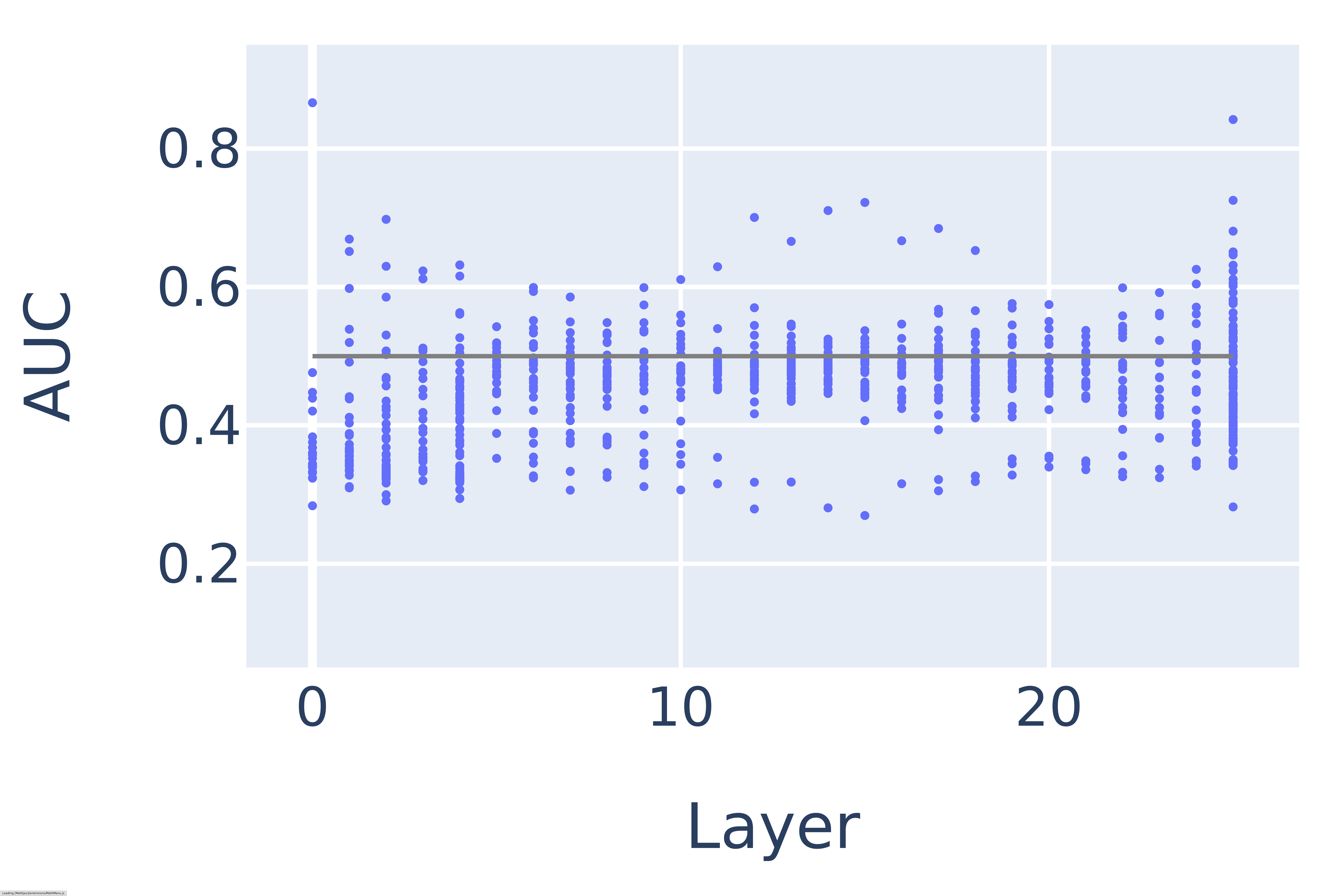}
    \includegraphics[width=0.24\textwidth]{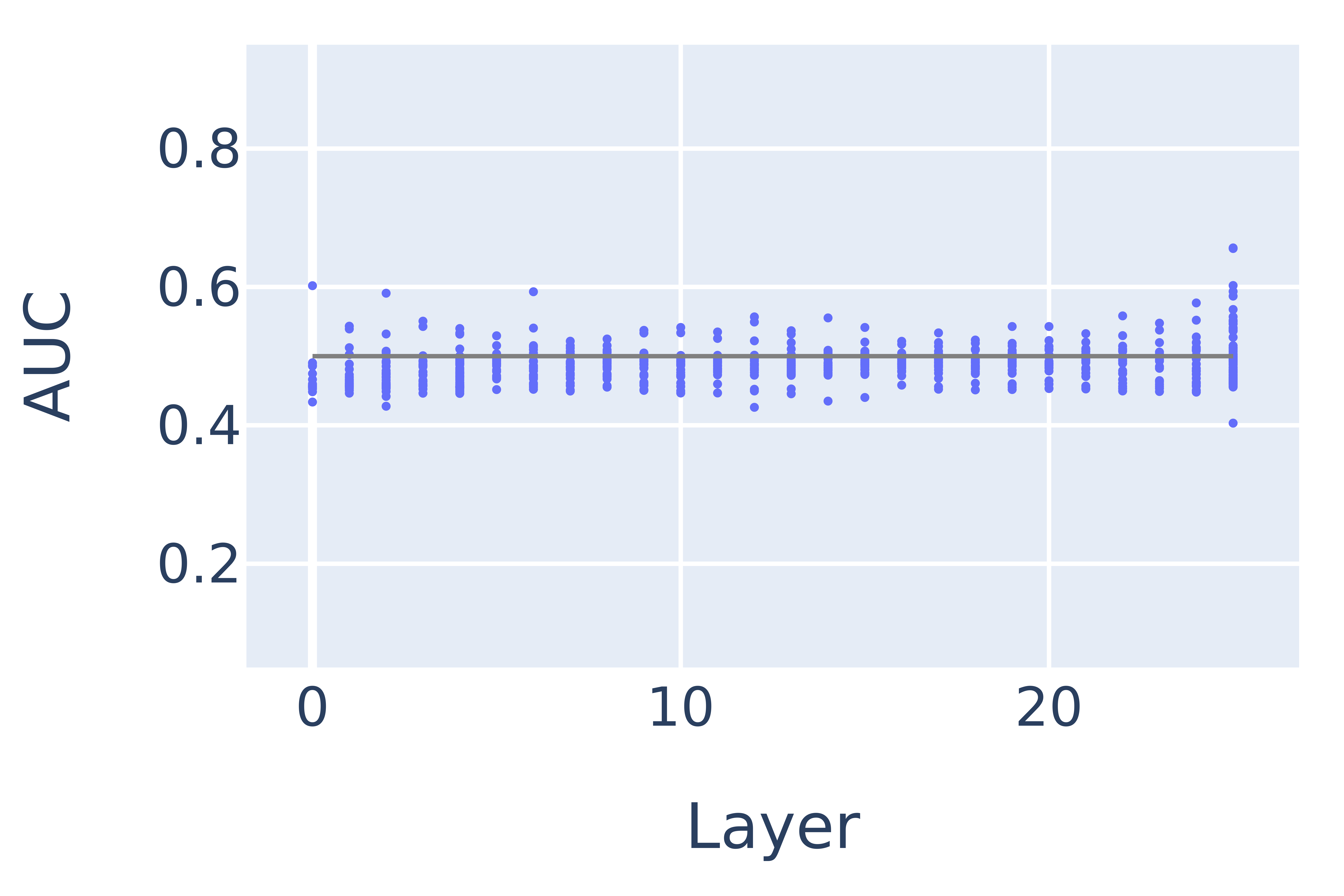}
    \includegraphics[width=0.24\textwidth]{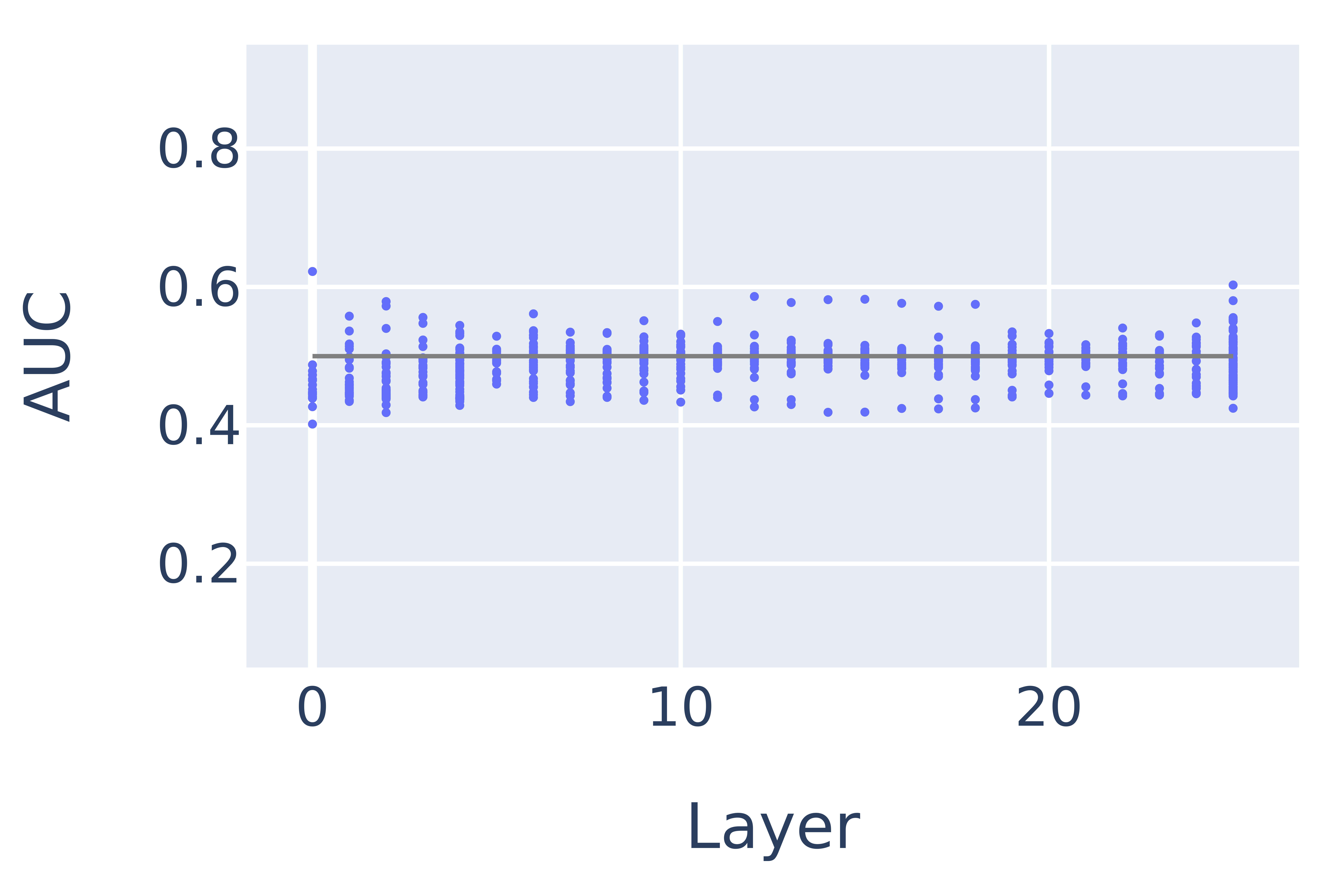}
    \includegraphics[width=0.24\textwidth]{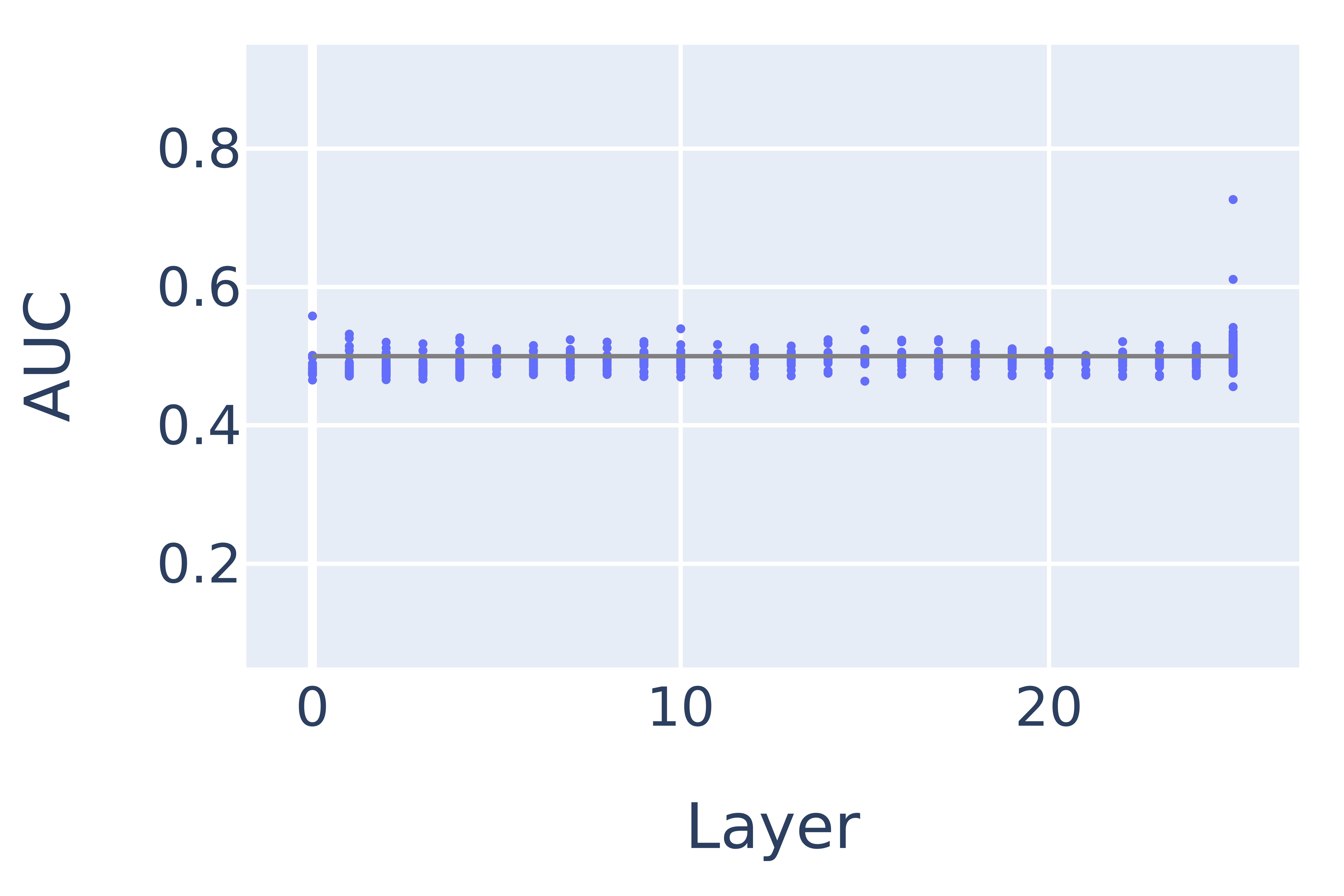}
\caption{From left to right, we show the AUCs of predicting latent firing using function words (any of \{'article', 'prepos', 'conjunction', 'det', 'modal', 'be', 'do', 'have', 'what'\}), articles, prepositions and conjunctions. These do not do as well as the ``meaningful-word'' or ``noun/propernoun'' groupings.
}
  \label{fig:additional_pos}
\end{figure}

\begin{table}[t]
  \centering
  \small
  \setlength{\tabcolsep}{4pt}  
  \begin{tabular}{ll}
    \toprule
    \textbf{Category} & \textbf{Tags} \\
    \midrule
    punc         & \texttt{. ( ) * -- , : `` '' \textquotesingle} \\
    quantifier   & \texttt{ABL, ABN, ABX, AP, AP\$} \\
    article      & \texttt{AT} \\
    be           & \texttt{BE, BED, BEDZ, BEG, BEM, BEN, BER, BEZ} \\
    conjunction  & \texttt{CC, CS} \\
    num          & \texttt{CD, OD} \\
    do           & \texttt{DO, DOD, DOZ} \\
    det          & \texttt{DT, DTI, DTS, DTX, DT\$} \\
    have         & \texttt{HV, HVD, HVG, HVN, HVZ} \\
    prepos       & \texttt{IN, TO} \\
    adj          & \texttt{JJ, JJR, JJS, JJT} \\
    modal        & \texttt{MD} \\
    noun         & \texttt{NN, NN\$, NNS, NNS\$, NR, NRS, NR\$, UH} \\
    propernoun   & \texttt{NP, NP\$, NPS, NPS\$} \\
    pronoun      & \texttt{PN, PN\$, PP\$, PP\$\$, PPL, PPLS, PPO, PPS, PPSS} \\
    qual         & \texttt{QL, QLP} \\
    adv          & \texttt{RB, RB\$, RBR, RBT, RN, RP} \\
    verb         & \texttt{VB, VBD, VBG, VBN, VBZ} \\
    what         & \texttt{WDT, WP\$, WPO, WPS, WQL, WRB, EX} \\
    unknown      & \texttt{NIL} \\
    \bottomrule
  \end{tabular}
  \vspace{1mm}
  \caption{Mapping from high-level category to Penn Treebank tags.  A trailing \texttt{\$} marks possessive forms.}
  \label{tab:ptb_map}
\end{table}

\subsection{Steering context-binding latents}
\label{app:steering_context_binding}
Our methodology for steering is as follows:
\begin{enumerate}
\item Prompt Gemma 2 2B with input text from the RedPajama dataset, ending at a natural point (after a newline token), with at least 400 tokens.
\item Capture the activating phrases of F1 and F2 that are at least 5 consecutive tokens long.
\item Allow Gemma 2 2B to generate a completion without steering, and prompt an LLM (Gemini 2.5 Flash Preview) to judge whether the completion is more like F1 activating examples, F2 activating examples, or unclear.
\item Repeat the above, but steering on the last token during generation, in the direction of F1 and F2. Since F1 and F2 are antipodal pairs, we first ablate the subspace spanned by F1 and F2, before adding the steering vector, that is fixed at 2x the historical activation of that feature in that context.

\end{enumerate}

\subsection{Meaningful-Word Latents}
\label{app:meaningful}

We provide the mapping from higher-level categories to Penn Treebank tags in \cref{tab:ptb_map}. In addition to the AUC for the categories shown in the text, we show the AUC for a few other categories in \cref{fig:additional_pos}, which seem mostly unrelated to predicting the activations of dense latents.

\subsection{PCA latents}
\label{app:pca}
\cref{fig:pc15} reports, layer-by-layer, each dense latent’s cosine similarity to PC1 (left) and the share of its norm contained in the top-5 PCs (right). As noted, only one antipodal pair per layer strongly aligns with PC1; most dense latents place little mass in the top PCs. \cref{fig:pcdiff} shows that this pattern is stable under SAE hyperparameters: varying the sparsity target ($L_0$) or dictionary size yields similar PC1 alignments, neither removing nor proliferating PC-aligned latents.

\begin{figure}[t]
    \centering
    \includegraphics[width=0.4\linewidth]{figures/pc1_percent_pca.pdf}
    \includegraphics[width=0.4\linewidth]{figures/pc5_percent_pca.pdf}
    \caption{(Left) Cosine similarity of dense latents with top principal component. (Right) Fraction norm of dense latents in top 5 principal components.}
    \label{fig:pc15}
\end{figure}

\begin{figure}[t]
    \centering
    \includegraphics[width=0.4\textwidth]{figures/pc1_cosine_similarity_diffL0.pdf}
    \includegraphics[width=0.4\textwidth]{figures/pc1_cosine_similarity_diffwidths.pdf}
\caption{Cosine similarity of dense latents in layer 12 with the top principal component, across different L0s and SAE dictionary sizes.}
  \label{fig:pcdiff}
\end{figure}

\begin{table}[H]
\centering
\resizebox{\columnwidth}{!}{%
\tiny
\begin{tabular}{@{} p{5cm} p{6cm} @{}}
\toprule
\multicolumn{2}{@{} c @{}}{\textbf{\footnotesize Dense latent L12:7541 ($f = 0.421$)}} \\
\midrule
\textbf{In-context} & \textbf{All-context} \\
\midrule
Numerical values or categories related to real estate listings, such as prices, property types, or number of beds & A proper noun or a specific concept/entity within a sentence \\
\midrule
Mentions of people, groups, or entities involved in a discussion or action & The beginning of a new clause or phrase, often following a comma, parenthesis, or other punctuation, or indicating a shift in topic or focus \\
\midrule
Mentions of Donna Brazile or political news and commentary, often critical of establishment figures or media outlets & A sequence of words that are part of a proper noun, a title, or a specific phrase, often capitalized, or a short phrase that acts as a label or identifier within a larger text. \\
\midrule
References to the Niagara Wire publication or its staff and content & The beginning of a new sentence or clause, often following punctuation or a line break \\
\midrule
Terms and phrases related to the Cybercrime Prevention Act of 2012 in the Philippines & The detection of numbers, dates, or specific numerical references within text \\
\midrule
References to Barack Obama or his administration & The beginning of a new clause or phrase, often following a conjunction, preposition, or punctuation mark, that introduces additional information or a new element into the sentence structure \\
\midrule
Mentions of specific people, places, or things, or references to a particular time or event & A proper noun or a common noun that is part of a larger phrase, often appearing after a preposition or as an object of a verb \\
\midrule
Mentions of CSIRO's internal operations, expertise, and collaborative efforts & Mentions of specific entities, objects, or concepts within a broader context, often highlighting a particular detail or aspect of the surrounding text \\
\midrule
Nautical vessel chartering and related services & The letter 'I' or 'A' or 'O' when it is the first letter of a word or a standalone word \\
\midrule
Mentions of healthcare organizations, roles, or initiatives related to health and human services & A token or sequence of tokens that is part of a larger, multi-word proper noun, compound noun, or specific phrase, where the preceding context helps to complete the meaning of the highlighted part. \\
\midrule
\multicolumn{2}{@{} c @{}}{\textbf{\footnotesize Dense latent L12:2009 ($f = 0.319$)}} \\
\midrule
\textbf{In-context} & \textbf{All-context} \\
\midrule
Punctuation marks, numbers, or single letters that are not part of a larger word & Code, symbols, or foreign language phrases \\
\midrule
Mentions of events, dates, or outcomes related to the life of Stéphanie of Monaco & The continuation of a word or phrase across a line break or other formatting boundary \\
\midrule
Mentions of the St. John's Red Storm basketball team, their coach Mike Jarvis, or their player Hatten & Short, common words or symbols that are often part of a larger phrase or structure, but do not carry significant meaning on their own \\
\midrule
Reporting on COVID-19 cases and related news & The beginning of a new sentence or clause, often following a period, comma, or other punctuation, or a line break \\
\midrule
Scientific citation formatting and punctuation & Punctuation marks, prepositions, and conjunctions that connect different parts of a sentence or list \\
\midrule
The beginning of a new clause or sentence & The beginning of a new word or token that is not preceded by a space \\
\midrule
References to students or pupils in an educational context & The detection of a word or phrase that is part of a larger, well-known entity or common expression, where the detected part is not the beginning of the entity or expression \\
\midrule
Biographical details and life events of a character, including family, career, and personal status, often presented in a chronological or list-like format & Mentions of specific people, places, or entities, or phrases that introduce or refer to them \\
\midrule
Specific references to the current or a past UN General Assembly session, or to the UN Secretary General and his staff & The beginning of a new sentence or phrase, often following punctuation or a line break, or the start of a new section within a document \\
\midrule
Mentions of specific dates, years, or numbers in a historical or official context & Mentions of specific words or phrases that are part of a larger list or enumeration, often found in titles, bullet points, or structured content \\
\bottomrule
\end{tabular}
}
\vspace{1mm}
\caption{Sampled explanations of dense latents L12:7541 and L12:2009, using in-context examples versus all-context examples. The in-context explanations are highly specific and diverse, while the all-context explanations are vague.}
\label{tab:contextbinding_dense_combined}
\end{table}

\begin{table}[H]
\centering
\resizebox{\columnwidth}{!}{%
\tiny
\begin{tabular}{@{} p{5cm} p{6cm} @{}}
\toprule
\multicolumn{2}{@{} c @{}}{\textbf{\footnotesize Sparse latent L12:10356 ($f=8.90 \times 10^{-4}$)}} \\
\midrule
\textbf{In-context} & \textbf{All-context} \\
\midrule
Years in the 2010s or 2020s, often following a movie title and sometimes preceded by "HBO" or "HBO Max" & A two-digit year following a month or day, or as part of a date range \\
\midrule
The number "1" in a four-digit year, specifically in the 2010s decade & The last two digits of a four-digit year \\
\midrule
The year 2019 in dates or as a standalone year & The last two digits of a four-digit year \\
\midrule
The digit '1' when it is part of a four-digit year that starts with '20' and is followed by a two-digit number, typically representing a day or a time, or a forward slash. & The second digit of a year in the 21st century \\
\midrule
The first digit of a two-digit year in a citation & The last two digits of a four-digit year \\
\midrule
The first two digits of a four-digit year & The last two digits of a year in the 21st century \\
\midrule
The release year of a movie title & The last two digits of a four-digit year \\
\midrule
Mentions of the "Product of the Year" award followed by a specific year & The last two digits of a year in a date \\
\midrule
The year 2022 in date formats & A four-digit year in the 21st century, specifically between 2010 and 2024 \\
\midrule
The third digit of a four-digit year, specifically when the year is 2013, often found in movie titles or release dates & The last two digits of a four-digit year, typically in the 2000s \\
\midrule
\multicolumn{2}{@{} c @{}}{\textbf{\footnotesize Sparse latent L12:800 ($f = 7.30 \times 10^{-4}$)}} \\
\midrule
\textbf{In-context} & \textbf{All-context} \\
\midrule
Mentions of the South Ossetian conflict, including locations, people, and related events & Mentions of people's names or titles, often followed by their statements or actions \\
\midrule
References to a specific person named Sarah, including possessive forms and direct address & Mentions of a person or entity speaking or being referenced \\
\midrule
Mentions of reports, analyses, or statements made by individuals or groups & A proper noun or pronoun that is the subject of a sentence or clause, or a proper noun that is the object of a verb or preposition \\
\midrule
Mentions of individuals or organizations involved in mine clearance or humanitarian aid & Mentions of people or organizations, often in attribution or as subjects of actions \\
\midrule
Mentions of the author Tom Bissell, often in relation to his work or statements & Mentions of proper nouns, often names of people or organizations, that are split across a line break \\
\midrule
Mentions of "Dicko" as a proper noun, often followed by a verb or punctuation, indicating a new clause or action related to the person. & Mentions of people speaking or being quoted \\
\midrule
Mentions of Alan Waller, Earl Spencer's former head of security & Proper nouns or pronouns referring to people or organizations, often followed by a verb \\
\midrule
Mentions of people or organizations as subjects or possessive entities & Mentions of a person's name followed by a verb of speaking or a reference to that person \\
\midrule
Attributions of statements or opinions to individuals or groups, often experts, in news or analytical contexts & A proper noun or pronoun that is the subject of a sentence or clause \\
\midrule
Mentions of Sonny Dykes, a football coach, or his last name, often in the context of his statements or actions & Mentions of people or organizations as subjects or agents of actions \\
\bottomrule
\end{tabular}
}
\vspace{1mm}
\caption{Sampled explanations of sparse latents L12:10356 and L2:800, using in-context and all-context examples. While the in-context explanations still tend to be more specific, they still center around a similar theme as the all-context explanations, and it is plausible that L12:10356 is a ``date'' feature and L12:800 is a ``proper noun'' feature.}
\label{tab:contextbinding_sparse}
\end{table}

\newpage
\pagebreak
\newpage
\newpage
\section*{NeurIPS Paper Checklist}

\begin{enumerate}

\item {\bf Claims}
    \item[] Question: Do the main claims made in the abstract and introduction accurately reflect the paper's contributions and scope?
    \item[] Answer:  \answerYes{} 
    \item[] Justification: Yes, we claim that we examine the geometry of high frequency latents, taxonomize high frequency latents, and examine their distribution across layers, all of which we do in the paper (we link to these specific sections in the introduction).

\item {\bf Limitations}
    \item[] Question: Does the paper discuss the limitations of the work performed by the authors?
    \item[] Answer: \answerYes{} 
    \item[] Justification: We have an extensive limitations section in the discussion section.

\item {\bf Theory assumptions and proofs}
    \item[] Question: For each theoretical result, does the paper provide the full set of assumptions and a complete (and correct) proof?
    \item[] Answer: \answerNA{}
    \item[] Justification: The paper does not include theoretical results.

    \item {\bf Experimental result reproducibility}
    \item[] Question: Does the paper fully disclose all the information needed to reproduce the main experimental results of the paper to the extent that it affects the main claims and/or conclusions of the paper (regardless of whether the code and data are provided or not)?
    \item[] Answer: \answerYes{} 
    \item[] Justification: We include experimental details sufficient to reproduce the experiments in the main body (all of which use open source models). We provide additional experiment details in \cref{app:exp_details}.

\item {\bf Open access to data and code}
    \item[] Question: Does the paper provide open access to the data and code, with sufficient instructions to faithfully reproduce the main experimental results, as described in supplemental material?
    \item[] Answer: \answerYes{} 
    \item[] Justification: We plan to upload our code by the supplementary materials deadline.

\item {\bf Experimental setting/details}
    \item[] Question: Does the paper specify all the training and test details (e.g., data splits, hyperparameters, how they were chosen, type of optimizer, etc.) necessary to understand the results?
    \item[] Justification: As described in the experimental details section, we include experimental details sufficient to reproduce the experiments in the main body (all of which use open source models). We provide additional experiment details in \cref{app:exp_details}.

\item {\bf Experiment statistical significance}
    \item[] Question: Does the paper report error bars suitably and correctly defined or other appropriate information about the statistical significance of the experiments?
    \item[] Answer: \answerNo{} 
    \item[] Justification: Because we are providing a taxonomy of different SAE latents, statistical significance in most of our experiments does not make sense.

\item {\bf Experiments compute resources}
    \item[] Question: For each experiment, does the paper provide sufficient information on the computer resources (type of compute workers, memory, time of execution) needed to reproduce the experiments?
    \item[] Answer: \answerYes{} 
    \item[] Justification: See \cref{app:compute_used} for a discussion of the compute used for our experiments.
    
\item {\bf Code of ethics}
    \item[] Question: Does the research conducted in the paper conform, in every respect, with the NeurIPS Code of Ethics \url{https://neurips.cc/public/EthicsGuidelines}?
    \item[] Answer: \answerYes{} 
    \item[] Justification: The research complies in every respect with the code of ethics.

\item {\bf Broader impacts}
    \item[] Question: Does the paper discuss both potential positive societal impacts and negative societal impacts of the work performed?
    \item[] Answer: \answerYes{} 
    \item[] Justification: See \cref{app:broader_impact} for a discussion of our paper's broader impact, which we overall believe to be very positive.
    
\item {\bf Safeguards}
    \item[] Question: Does the paper describe safeguards that have been put in place for responsible release of data or models that have a high risk for misuse (e.g., pretrained language models, image generators, or scraped datasets)?
    \item[] Answer: \answerNA{} 
    \item[] Justification: We do not release any new data or models.

\item {\bf Licenses for existing assets}
    \item[] Question: Are the creators or original owners of assets (e.g., code, data, models), used in the paper, properly credited and are the license and terms of use explicitly mentioned and properly respected?
    \item[] Answer: \answerYes{} 
    \item[] Justification: We use and cite the open source Gemma models and GemmaScope SAEs.

\item {\bf New assets}
    \item[] Question: Are new assets introduced in the paper well documented and is the documentation provided alongside the assets?
    \item[] Answer: \answerNA{} 
    \item[] Justification: We do not introduce any new assets.

\item {\bf Crowdsourcing and research with human subjects}
    \item[] Question: For crowdsourcing experiments and research with human subjects, does the paper include the full text of instructions given to participants and screenshots, if applicable, as well as details about compensation (if any)? 
    \item[] Answer: \answerNA{} 
    \item[] Justification: We do not do research with human subjects.

\item {\bf Institutional review board (IRB) approvals or equivalent for research with human subjects}
    \item[] Question: Does the paper describe potential risks incurred by study participants, whether such risks were disclosed to the subjects, and whether Institutional Review Board (IRB) approvals (or an equivalent approval/review based on the requirements of your country or institution) were obtained?
    \item[] Answer: \answerNA{} 
    \item[] Justification: We do not do research with human subjects.

\item {\bf Declaration of LLM usage}
    \item[] Question: Does the paper describe the usage of LLMs if it is an important, original, or non-standard component of the core methods in this research? Note that if the LLM is used only for writing, editing, or formatting purposes and does not impact the core methodology, scientific rigorousness, or originality of the research, declaration is not required.
    \item[] Answer: \answerNA{}
    \item[] Justification: The core method development in this research does not involve LLMs as any important, original, or non-standard components.

\end{enumerate}

\end{document}